\documentclass{article}

\PassOptionsToPackage{numbers, compress}{natbib}

\usepackage[final]{neurips_2023}

\usepackage[utf8]{inputenc} 
\usepackage[T1]{fontenc}    
\usepackage{hyperref}       
\usepackage{hyperref}
\hypersetup{
    colorlinks=true,
    linkcolor=[rgb]{0.0, 0.2, 0.5},
    urlcolor=[rgb]{0.0, 0.2, 0.5},
    filecolor=[rgb]{0.0, 0.2, 0.5},
    citecolor=cyan,
    pdftitle={My document},
    pdfpagemode=FullScreen,
    pdfborder={0 0 0},
    pdfhighlight=/N
}
\usepackage{url}            
\usepackage{booktabs}       
\usepackage{amsfonts}       
\usepackage{nicefrac}       
\usepackage{microtype}      
\usepackage{xcolor}         
\usepackage{graphicx}       
\usepackage{wrapfig}        
\usepackage{amsmath}        
\usepackage{amssymb}
\usepackage{multirow}       
\usepackage{multicol}
\usepackage{algorithm}      
\usepackage{algpseudocode}
\usepackage{appendix}       
\usepackage{bm}

\title{DiffSketcher: Text Guided Vector Sketch Synthesis through Latent Diffusion Models}

\author{%
  Ximing Xing \\
  Beihang University\\
  \texttt{ximingxing@buaa.edu.cn} \\
  \And
  Chuang Wang \\
  Beihang University\\
  \texttt{chuangwang@buaa.edu.cn} \\
  \And
  Haitao Zhou \\
  Beihang University\\
  \texttt{zhouhaitao@buaa.edu.cn} \\
  \And
  Jing Zhang \\
  Beihang University\\
  \texttt{zhang\_jing@buaa.edu.cn} \\
  \And
  Qian Yu\thanks{Corresponding author} \\
  Beihang University\\
  \texttt{qianyu@buaa.edu.cn} \\
  \And
  Dong Xu \\
  The University of Hong Kong\\
  \texttt{dongxu@cs.hku.hk} \\
}

\author{%
  Ximing Xing$^{1}$,
  Chuang Wang$^{1}$,
  Haitao Zhou$^{1}$,
  Jing Zhang$^{1}$,
  Qian Yu$^{1}$\thanks{Corresponding author},
  Dong Xu$^{2}$ \\
  $^{1}$Beihang University \quad $^{2}$The University of Hong Kong \\
  \texttt{\{ximingxing, qianyu\}@buaa.edu.cn} \quad
  \texttt{dongxu@cs.hku.hk}
}

\begin{document}

\maketitle

\begin{figure*}[h]
\centering
\includegraphics[width=0.95\linewidth]{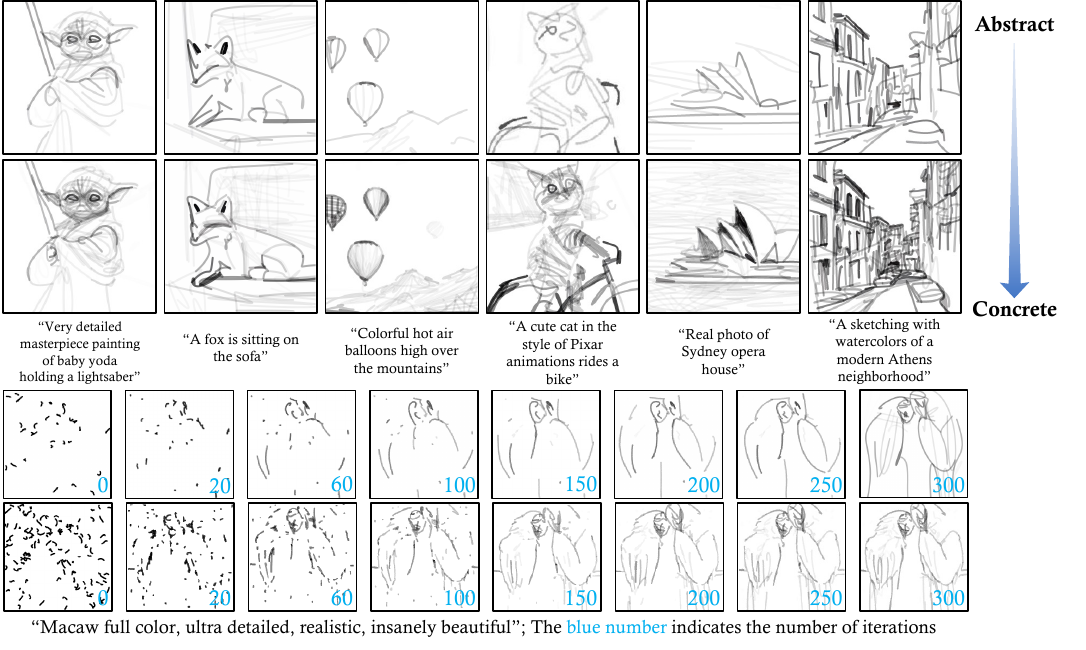}
\vspace{-1em}
\caption{
\textbf{Capabilities of DiffSketcher.} Our method enables controllable text-to-sketch generation. \textbf{Top:} By adjusting the number of strokes, we can synthesize sketches at various abstraction levels (abstract vs. concrete). \textbf{Bottom:} The progressive drawing process of a generated sketch.
}
\label{fig:tears}
\end{figure*}

\begin{abstract}
We demonstrate that pre-trained text-to-image diffusion models, despite being trained on raster images, possess a remarkable capacity to guide vector sketch synthesis. In this paper, we introduce \textit{DiffSketcher}, a novel algorithm for generating \textit{vectorized} free-hand sketches directly from natural language prompts. Our method optimizes a set of Bézier curves via an extended Score Distillation Sampling (SDS) loss, successfully bridging a raster-level diffusion prior with a parametric vector generator. To further accelerate the generation process, we propose a stroke initialization strategy driven by the diffusion model's intrinsic attention maps. Results show that \textit{DiffSketcher} produces sketches across varying levels of abstraction while maintaining the structural integrity and essential visual details of the subject. Experiments confirm that our approach yields superior perceptual quality and controllability over existing methods. The code and demo are available at \href{https://ximinng.github.io/DiffSketcher-project/}{https://ximinng.github.io/DiffSketcher-project/}.
\end{abstract}

\section{Introduction}
Minimal representations, such as natural language and free-hand sketches, are powerful tools for effectively conveying ideas by emphasizing the essence of a subject. While natural language encapsulates abstract semantics, sketches provide a visual abstraction that makes these concepts more concrete. For instance, designers often sketch prototypes based on a client's verbal description to ensure mutual understanding. Automating this text-to-sketch translation process could significantly streamline design workflows and reduce manual effort.

Despite its vast potential, the task of direct text-to-sketch generation remains largely unexplored. Existing research has primarily focused on image-conditioned sketch synthesis~\cite{infodrawing_Chan_2022,liu2020unsupervised,vinker_2022_clipasso,clipascene_vinker_2022}. For example, Info-drawing~\cite{infodrawing_Chan_2022} employs a generative adversarial network (GAN) to generate raster sketches from input photos, utilizing a style loss to mimic reference sketches. CLIPasso~\cite{vinker_2022_clipasso} introduces a differentiable rasterizer optimized via semantic loss to produce vector sketches, which its follow-up work, CLIPascene~\cite{clipascene_vinker_2022}, extends to scene-level images. However, these methods inherently rely on visual reference images and cannot generate fundamentally \textit{new content} directly from textual descriptions.

\begin{figure}[p]
\centering
\includegraphics[width=\textwidth,height=\textheight,keepaspectratio]{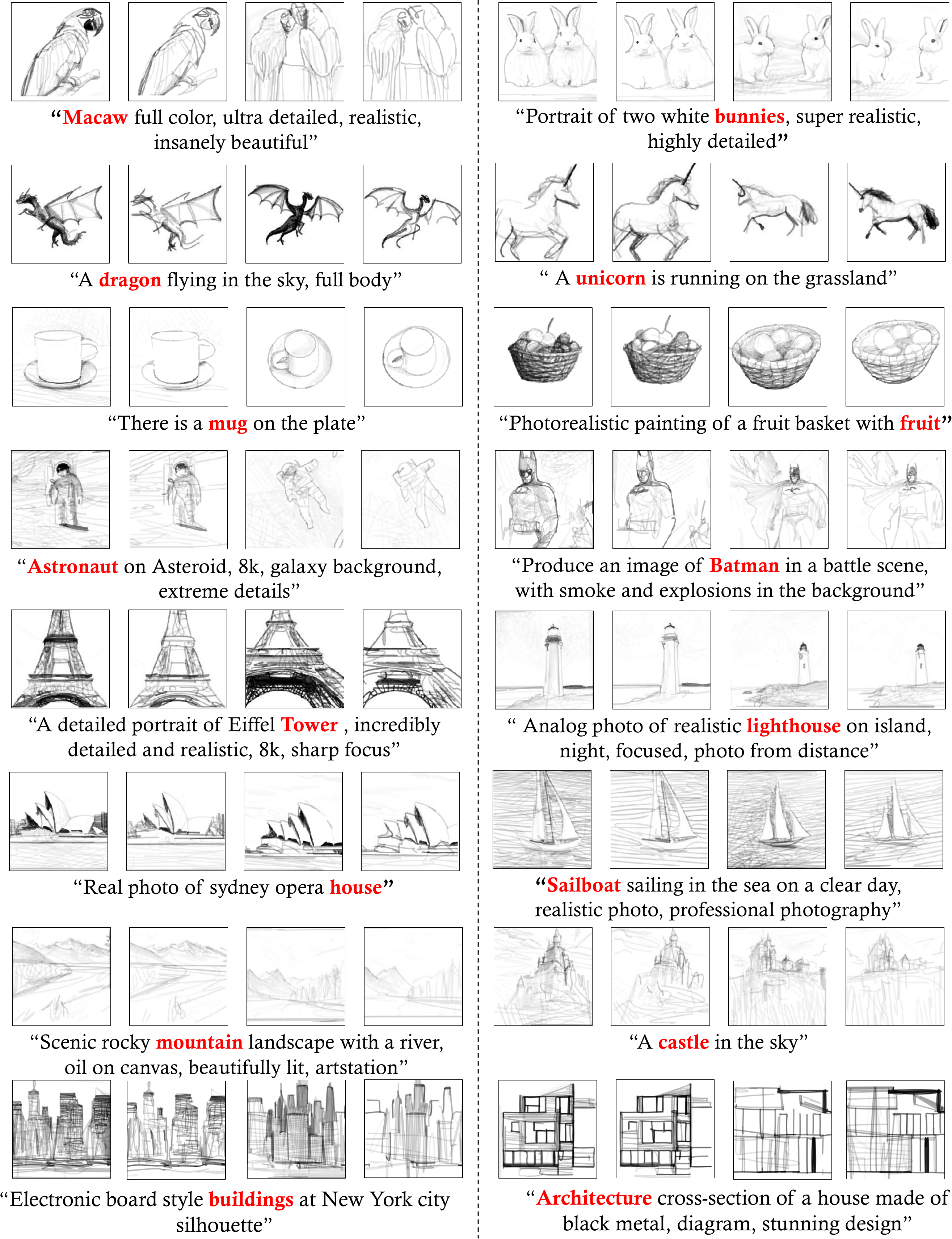}
\vspace{-1em}
\caption{
\textbf{Various free-hand sketches synthesized by DiffSketcher.}
DiffSketcher obtains prior information from LDM~\cite{ldm_2022_Rombach} composite images through score distillation~\cite{poole_2023_dreamfusion} and achieves the same heavy and light drawing styles as human sketches by performing gradient descent on a set of Bézier curves with the opacity property.
Our proposed DiffSketcher allows for varying levels of abstraction while matching its corresponding textual semantics.
In each example, given the same text prompt and two different random seeds, two sketches with a different number of strokes are generated. The red words represent the cross-attention index used to initialize the control points (details about cross-attention are covered in Section~\ref{sec:stroke_init}).
}
\label{fig:more_results}
\vspace{-1.5em}
\end{figure}

Concurrently, text-to-image generation has witnessed remarkable breakthroughs driven by diffusion models~\cite{GLIDE_2022_nichol,DALLE2_2022_ramesh,ldm_2022_Rombach,imagen_2022_saharia} trained on massive image-text datasets~\cite{laion_schuhmann_2022}. While these models excel at high-fidelity, controllable image synthesis, they struggle to natively produce highly abstract and \textit{vectorized} free-hand sketches (as shown in Fig.~\ref{fig:more_results}). Inspired by the success of text-to-image diffusion models and image-conditioned sketch generators, we aim to build a bridge between these two fundamental forms of human expression.

In this work, we present \textit{DiffSketcher}, a novel algorithm capable of synthesizing high-quality, free-hand vector sketches directly from natural language prompts. Crucially, \textit{DiffSketcher} operates without the need for text-sketch paired data or large sketch datasets. Instead, we define a sketch as a set of parametric Bézier curves and optimize them through a differentiable rasterizer~\cite{Li_2020_diffvg}, guided by a pre-trained text-to-image diffusion model~\cite{ldm_2022_Rombach}. The core idea is to distill the rich prior knowledge of a raster-level diffusion model into the vector optimization process. This ensures that the generated sketches are semantically coherent and strictly aligned with the text prompts. However, fully harnessing this diffusion prior to efficiently generate both simple objects and complex scenes poses a non-trivial challenge.

To address this, we propose three key strategies to enhance generation quality and efficiency: 
(1) \textbf{Extended Score Distillation Sampling (SDS) Loss:} While previous works rely heavily on CLIP loss to optimize vector parameters, we introduce an extended SDS loss that yields significantly more diverse sketch synthesis results. It can also be seamlessly combined with CLIP or LPIPS losses to provide supplementary control.
(2) \textbf{Attention-Driven Stroke Initialization:} Randomly initializing stroke starting points makes the synthesis process highly time-consuming. By fusing the cross-attention and self-attention maps from the diffusion model's U-Net~\cite{Dhariwal_ADM_2021}, we establish a highly effective stroke initialization strategy that drastically accelerates convergence compared to random initialization.
(3) \textbf{Stroke Opacity Optimization:} We introduce an opacity property during the optimization of the Bézier curves to simulate the varying pressure and dynamic brushstrokes of human drawing, achieving a more natural and expressive visual effect.

In summary, our main contributions are three-fold: 
(1) We propose \textit{DiffSketcher}, a pioneering text-to-sketch generation framework. To the best of our knowledge, it is the first method capable of generating diverse, high-quality vector sketches at various levels of abstraction for both object and scene levels, without relying on sketch datasets. 
(2) We design a robust optimization pipeline featuring an extended SDS loss, an attention-guided stroke initialization strategy, and opacity modeling, which collectively enhance generation efficiency and structural accuracy. 
(3) We conduct extensive experiments demonstrating that \textit{DiffSketcher} significantly outperforms existing alternatives in both visual quality and semantic alignment, establishing a strong baseline for future research.

\section{Related Work}
\noindent\textbf{Sketch Synthesis.}\quad
Free-hand drawings convey abstract concepts through human visual perception with minimal abstraction. Unlike purely edge-map extraction methods~\cite{canny_1986}, free-hand sketching aims to present sketches that are abstract in terms of structure~\cite{infodrawing_Chan_2022} and semantic interpretation\cite{vinker_2022_clipasso}. Therefore, computational sketching methods that aim to mimic human drawing consider a wide range of sketch representations, ranging from those grounded in the edge map of the input image\cite{xie_2015_holistically,free_li_2017, li2019photo, grayscale_tong_2021,liu2020unsupervised,deepfacedrawing_chen_2020} to those that are more abstract~\cite{sketchRNN_ha_2018a,creative_ge_2021,doodleformer_bhunia_2022, clipdraw_frans_2022, SketchLattice_Qi_2021,vinker_2022_clipasso}, which are normally in vector format. Among the works synthesizing vector sketches, CLIPasso~\cite{vinker_2022_clipasso} and CLIPascene~\cite{clipascene_vinker_2022} are conditioned on an input image, while the rest are unconditional. Until now, no prior work has explored synthesizing a sketch based on text. 

\noindent\textbf{Vector Graphics.}\quad Our work builds upon the differentiable renderer for vector graphics introduced by Li et al~\cite{Li_2020_diffvg}. While image generation methods that operate over vector images traditionally require a vector-based dataset, recent work has shown how the differentiable renderer can be used to bypass this limitation~\cite{clipgen_shen_2021,rethinking_kotovenko_2021, evolution_tian_2022, im2vec_reddy_2021, Ma_2022_LIVE}. Furthermore, recent advances in visual text embedding contrastive language-image pre-training (CLIP)\cite{clip_radford_2021} have enabled a number of successful methods for synthesizing sketches, such as CLIPDraw~\cite{clipdraw_frans_2022}, StyleCLIPDraw~\cite{Styleclipdraw_schaldenbr_2021}, CLIP-CLOP~\cite{clipclop_mirowski_2022}, and CliPascene~\cite{clipascene_vinker_2022}. A very recent work VectorFusion~\cite{jain2022vectorfusion} combine differentiable renderer with diffusion model for vector graphics generation, \textit{e.g.}, iconography and pixel art. Our proposed algorithm, DiffSketcher, shares a similar idea with VectorFusion, but our focus is generating object- and scene-level sketches from a natural language prompt.

\noindent\textbf{Diffusion Models.} Denoising diffusion probabilistic models (DDPMs)~\cite{diffusion_models_dickstein_2015,EestGrad_song_2019,ddpm_ho_2020,scorebased_song_2021}, particularly those conditioned on text, have shown promising results in text-to-image synthesis. For example, Classifier-Free Guidance (CFG)~\cite{classifierfree_2022_ho} has improved sample quality and has been widely used in large-scale diffusion model frameworks, including GLIDE~\cite{GLIDE_2022_nichol}, Stable Diffusion~\cite{ldm_2022_Rombach}, DALL·E 2~\cite{DALLE2_2022_ramesh}, and Imagen~\cite{imagen_2022_saharia}. However, the majority of images available in web-scale datasets are rasterized, and this work follows the framework of \textit{synthesis through optimization}, in which images are generated through evaluation-time optimization against a given metric. Our proposed algorithm, \textit{DiffSketcher}, uses a pre-trained text-to-image diffusion model to synthesize free-hand sketches from natural language input. This is achieved by transferring image synthesis prior information into a differentiable renderer.xc

\section{Preliminaries}

\subsection{Diffusion Models}

In this section, we provide a concise overview of diffusion models, which are a class of generative models that utilize latent variables to gradually transform a sample from a noise distribution to a target data distribution~\cite{diffusion_models_dickstein_2015,ddpm_ho_2020}.
Diffusion models consist of two components: a forward process $q$ and a reverse process or generative model $p$. The forward process, which is typically modeled as a Gaussian distribution, gradually removes structure from the input data $\bm{x}$ by adding noise over time. The reverse process, on the other hand, adds structure to the noise starting from a latent variable $\bm{z}_t$.
Specifically, the generative model is trained to slowly add structure starting from random noise $p(\bm{z}_T) = \mathcal{N}(\bm{0}, \bm{I})$ with transitions $p_{\phi}(\bm{z}_{t-1}|\bm{z}_{t})$.
$q(\bm{z}_t | \bm{x}) = \mathcal{N}(\alpha_t \bm{x}, \sigma^2_t \bm{I})$.
Training the diffusion model with a (weighted) evidence lower bound (ELBO) simplifies to a weighted denoising score matching objective for parameters $\phi$~\cite{ddpm_ho_2020}
\begin{equation}
\mathcal{L}_{\mathrm{Diff}}(\phi, \bm{x})=\mathbb{E}_{t \sim \mathcal{U}(0,1), \epsilon \sim \mathcal{N}(\bm{0}, \bm{I})}
\left[
w(t)\left\|\epsilon_{\phi}\left(\alpha_{t} \bm{x}+\sigma_{t} \epsilon ; t\right)-\epsilon\right\|_{2}^{2}
\right]
\end{equation}
where $w(t)$ is a weighting function that depends on the timestep $t$.
Our work builds on text-to-image latent diffusion model (LDM) that learn $\epsilon_{\phi}(\bm{z}_t;t,y)$ conditioned on text embeddings $y$~\cite{ldm_2022_Rombach}.
LDM uses classifier-free guidance(CFG)~\cite{classifierfree_2022_ho}, which jointly learns an unconditional model to enable higher quality generation via a guidance scale parameter $\omega: \hat{\epsilon}_{\phi} (\bm{z}_t;y,t) = (1+w) \epsilon_{\phi}(\bm{z}_t;y,t) - w\epsilon_{\phi}(\bm{z}_t;t)$ ($\hat{\epsilon}_{\phi}$ denotes the guided version of the noise prediction). CFG alters the score function to prefer regions where the ratio of the conditional density to the unconditional density is large. In practice, setting $w > 0$ improves sample fidelity at the cost of diversity.

\subsection{Score Distillation Sampling}

Many 3D generative approaches use a frozen image-text joint embedding model (\textit{e.g.} CLIP) and an optimization-based approach to train a Neural Radiance Fields (NeRF)~\cite{nerf_mildenhall_2021}.
Such models can be specified as a differentiable image parameterization (DIP)~\cite{differentiable_mordvintsev_2018}, where a differentiable generator $g$ transforms parameters $\theta$ to create an image $\bm{x} = g(\theta)$.
In DreamFusion~\cite{poole_2023_dreamfusion}, $\theta$ be parameters of a 3D volume and $g$ is a volumetric renderer.
To learn these parameters, DreamFusion proposed \textit{score distillation sampling} (SDS) loss that can be applied to Imagen~\cite{imagen_2022_saharia}:
\begin{equation}
\nabla_{\theta} \mathcal{L}_{\mathrm{SDS}} (\phi, \bm{x}=g(\theta)) \triangleq \mathbb{E}_{t,\epsilon} \left[ 
\omega(t) (\hat{\epsilon}_{\phi}(\bm{z};y,t) - \epsilon) 
\frac{\partial \bm{x}}{\partial \theta}
\right]
\end{equation}
where the constant $\alpha_t \bm{I} = \partial \bm{z}_t / \partial \bm{x}$ is absorbed into $w(t)$, and the classifier-free-guided $\hat{\epsilon}_{\phi}$ is used.
In practice, SDS gives access to loss gradients, not a scalar loss. Their proposed SDS loss provides a way to assess the similarity between an image and a caption:
\begin{equation}
\nabla_{\theta} \mathcal{L}_{\mathrm{SDS}}(\phi, \bm{x}=g(\theta))
=
\nabla_{\theta} \mathbb{E}_{t} \left[
\sigma_{t} / \alpha_{t} w(t) 
\mathrm{KL}
\left(
q\left(\bm{z}_{t} | g(\theta) ; y, t\right)|p_{\phi}\left(\bm{z}_{t} ; y, t\right)
\right)
\right]
\end{equation}
where $p_{\phi}$ is the distribution learned by the frozen, pretrained Imagen model. $q$ is a unimodal Gaussian distribution centered at a learned mean image $g(\theta)$.
DreamFusion~\cite{poole_2023_dreamfusion} proposed an approach to use a pretrained pixel-space text-to-image diffusion model (Imagen~\cite{imagen_2022_saharia}) as a loss function. However, diffusion models trained on pixels have traditionally been used to sample only pixels.
We want to create what look like good sketches that match the text prompts when rendered from a set of vector strokes.
Such models can be specified as a differentiable image parameterization, where a differentiable rasterizer $\mathcal{R}$ transforms parameters $\theta$ to create a sketch $\mathcal{S} = \mathcal{R}(\theta)$.

Inspired by DreamFusion~\cite{poole_2023_dreamfusion}, we extend \textit{score distillation sampling} (SDS) loss to use a pretrained latent diffusion model~\cite{ldm_2022_Rombach} as a prior for optimizing curve parameters.
Intuitively, score distillation converts diffusion sampling into an optimization problem that allows the raster image to be represented by a differentiable rasterizer.

\subsection{Differentiable Rasterizer}

Li \textit{et al.}~\cite{Li_2020_diffvg} introduce a differentiable rasterizer $\mathcal{R}$ that bridges the vector graphics and raster image domains. 
A raster image is a 2D grid sampling over the space of the vector graphics scene $f(x,y;\Theta)$, where $\Theta$ contains the curve parameters, \textit{e.g.,} coordinates of Bézier control points, opacity, line thickness. Given a 2D location $(x,y) \in \mathbb{R}^2$, they first find all the filled curves and strokes that overlap with the location. Then sort them with a user-specified order and compute the color using alpha blending~\cite{digital_images_porter_1984}. It enables the gradients of the rasterized curves (image) to be backpropagated to the curve parameters.

\section{Methodology}
\label{sec:method}

In this section, we present our method for generating sketches using pre-trained text-to-image diffusion models.
Let $\mathcal{S}$ be a sketch that was rendered by a differentiable rendering~\cite{Li_2020_diffvg} $\mathcal{R}$ using the text prompt $\mathcal{P}$. Our goal is to optimize a set of parametric strokes to automatically generate a vector sketch that matches the description of the text prompt.

\begin{wrapfigure}{r}{0.5\textwidth}
\vspace{-1em}
\centering
\includegraphics[width=0.3\textheight]{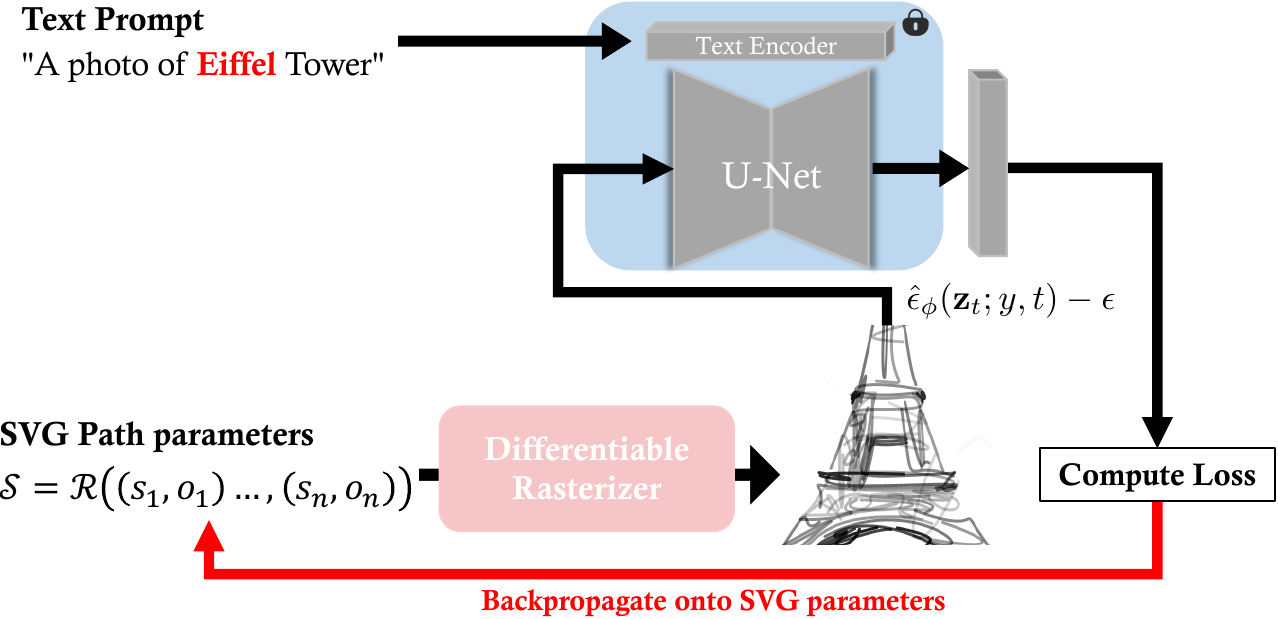}
\vspace{-0.8em}
\caption{
\footnotesize \textbf{The overview of the pipeline.} DiffSketcher accepts a set of control points (the locations of the strokes) and text prompts as input to generate a hand-drawn sketch.
} \label{fig:pipeline}
\vspace{-1.5em}
\end{wrapfigure}
Our pipeline is illustrated in Figure~\ref{fig:pipeline}. Given a text prompt $y$ of the desired subject and an initial set of stroke control points, we synthesize the corresponding sketch $\mathcal{S}$ while matching the semantic attributes of the prompt. To initialize these control points, we extract and fuse the attention maps from the U-Net~\cite{Dhariwal_ADM_2021} of the Latent Diffusion Model (LDM)~\cite{ldm_2022_Rombach}. Our key observation is that the structure and appearance of the generated image depend on the interaction between spatial latent features and text embeddings during the diffusion process~\cite{prompt2prompt_hertz_2023}. We provide more details on stroke initialization in Section~\ref{sec:stroke_init}.

\begin{figure*}[t]
\vspace{-1em}
\centering
\includegraphics[width=1\linewidth]{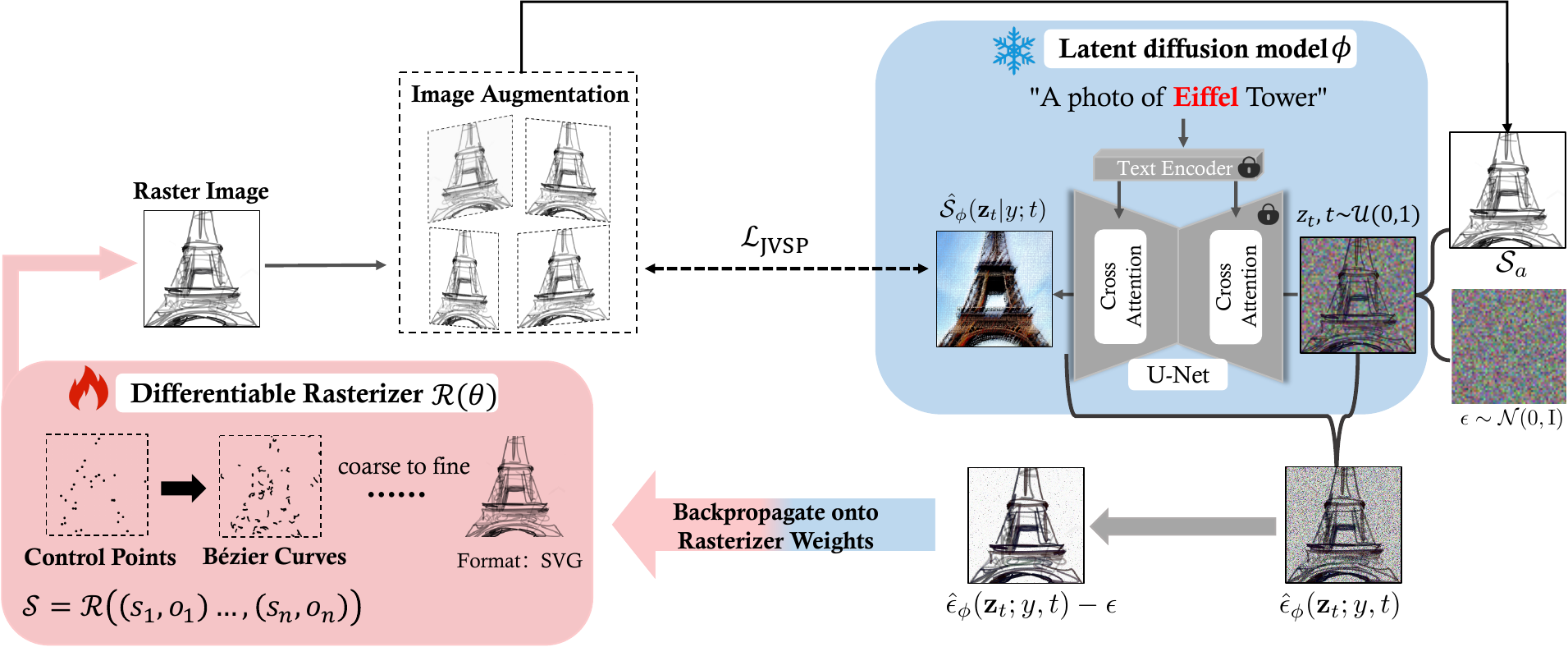}
\vspace{-1.5em}
\caption{
\textbf{Optimization overview.} 
To synthesize a sketch that matches the given text prompt, we optimize the parameters of the differentiable rasterizer $\mathcal{R}$ that produces the raster sketch $\mathcal{S}$, such that the resulting sketch is close to a sample from the frozen latent diffusion model (the blue part of the picture). 
Since the diffusion model directly predicts the update direction, we do not need to backpropagate through the diffusion model; the model simply acts like an efficient, frozen critic that predicts image-space edits.
} \label{fig:overview}
\vspace{-1em}
\end{figure*}

As shown in Figure~\ref{fig:overview}, at each optimization step, we feed the stroke parameters $\theta$ to a differentiable rasterizer $\mathcal{R}$ to produce the raster sketch $\mathcal{S} = \mathcal{R}(\theta)$. We optimize $\theta$ such that $\mathcal{S}$ aligns with the prior of the frozen LDM. To perform this optimization, we propose a variation of the SDS loss function that enhances the texture of hand-sketched strokes and improves the resilience of drawing styles, ensuring the raster sketch remains semantically coherent with the text prompt. The generated sketch, along with representations from the frozen LDM, is then used to formulate a joint semantic and perceptual loss. Finally, we backpropagate this loss through the differentiable rasterizer $\mathcal{R}$, updating the control points and stroke opacities directly at each step until convergence.

\subsection{Synthesis Through Optimization}
We define a sketch as a set of $n$ strokes $\{s_1, \dots, s_n\}$ with the opacity attribute placed on a white background. 
To represent each stroke, we use a two-dimensional Bézier curve with four control points $\mathbf{s}_i = \{p_i^j\}_{j=1}^4 = \{(x_i,y_i)^j\}_{j=1}^4$ (Notice that here, $x$ and $y$ represent the coordinates in the canvas) and one opacity attribute $o_i$.
We incorporate the opacity of the strokes into the optimization process and use DiffSketcher semantics understanding to achieve a human-like \textit{heavy and light sketch style}.
The parameters of the strokes are fed to a differentiable rasterizer $\mathcal{R}$, which forms the raster sketch $\mathcal{S} = \mathcal{R} ((s_1,o_1) \dots, (s_n, o_n)) = \mathcal{R} ((\{p_1^j\}_{j=1}^4, o_1), \dots, (\{p_n^j\}_{j=1}^4, o_n))$. For simplicity, we define the parameter in $\mathcal{R}$ as $\theta$.

\subsubsection{Vanilla Version: Fidelity to Generated Image}
\label{sec:vanilla_version}
We start with a two-stage pipeline: First, we sample an image from the latent diffusion model~\cite{ldm_2022_Rombach} using a text prompt. Next, we optimize the control points to obtain a sketch that is consistent with the text prompt. 
To preserve the fidelity of the generated sample, we incorporate a \textbf{J}oint \textbf{V}isual \textbf{S}emantic and \textbf{P}erceptual (JVSP) loss to optimize the similarity of the synthesized sketches and the instance sampled from the frozen latent diffusion model~\cite{ldm_2022_Rombach}.
We leverage the VAE~\cite{taming_2021_esser} decoder $\mathcal{D}$ to get the RGB pixel representation of $\hat{\mathcal{S}}_{\phi}(\bm{z}_t|y;t)$.
Then, we jointly use the LPIPS~\cite{lpips_zhang_2018} and CLIP visual encoders~\cite{clip_radford_2021,vinker_2022_clipasso} as depth structural similarity and visual semantic similarity metrics, respectively.
Specifically, we use the following loss function:
\begin{equation}
\mathcal{L}_{\mathrm{JVSP}} = 
\mathcal{L}_{\mathrm{LPIPS}}(\mathcal{D}(\hat{\mathcal{S}}_{\phi}(\bm{z}_t|y;t)), \mathcal{R}(\theta)) + 
\sum_l \left\| \mathrm{CLIP}_l (\mathcal{D}(\hat{\mathcal{S}}_{\phi}(\bm{z}_t|y;t))) - \mathrm{CLIP}_l (\mathcal{R}(\theta)) \right\|
\end{equation}
The $\mathcal{L}_{\mathrm{JVSP}}$ loss function encourages the synthesized sketch to match the underlying semantics and perceptual details of the image sampled from LDM, leading to more realistic and visually appealing results.
While vectorizing a rasterized diffusion sample is lossy, an input \textbf{A}ugmentation version of the \textbf{SDS} (ASDS) loss can either finetune the results or optimize random control points from scratch to sample a sketch that is consistent with the text prompt.
In the Sec~\ref{sec:ASDS_loss}, we introduce the ASDS loss to match the text prompt.

\subsubsection{Augmentation SDS Loss: Fidelity to Text Prompt}
\label{sec:ASDS_loss}
To synthesize a vector sketch that matches a given text prompt, we directly optimize the parameters $\theta$ of the differentiable rasterizer $\mathcal{R}$ that produces the raster sketch $\mathcal{S}$.
We propose an input augmentation version of the SDS loss function to perform this optimization, which encourages plausible images to have low loss and implausible images to have a high loss. 
Given a raster sketch $\tilde{\mathcal{S}} \in \mathbb{R}^{H \times W \times 3}$, we combine \textit{RandomPerspective}, \textit{RandomResizedCrop} and \textit{RandomAdjustSharpness} to get a data augmentation version of $\bm{\tilde{\mathcal{S}}_a} \in \mathbb{R}^{512 \times 512 \times 3}$. These transformations preserve artistic styles, enhance style diversity, emphasize crucial features, and improve the model's adaptability to various artistic renditions and size variations commonly found in black-and-white hand-drawn sketches.
Then, the LDM uses a VAE encoder~\cite{taming_2021_esser} to encode $\bm{\tilde{\mathcal{S}}_a}$ into a latent representation $\bm{z} = \mathcal{E}(\bm{\tilde{\mathcal{S}}}_a)$, where $\bm{z} \in \mathbb{R}^{(H / f) \times (W / f) \times 4}$ and $f$ is the encoder downsample factor.
In summary, we use the following ASDS loss function:
\begin{equation}
\nabla_{\theta} \mathcal{L}_{\mathrm{ASDS}} (\phi, \mathcal{S} = \mathcal{R}(\theta)) \triangleq 
\mathbb{E}_{t,\epsilon,a} 
\left[ 
w(t) (\hat{\epsilon}_{\phi}(\bm{z}_t;y,t) - \epsilon) 
\frac{\partial \bm{z}}{\partial \tilde{\mathcal{S}}_a}
\frac{\partial \tilde{\mathcal{S}}_a}{\partial \theta}
\right]
\end{equation}
Where the weighting function $w(t)$ is a hyper-parameter. And we sample $t \sim \mathcal{U}(0.05, 0.95)$, avoiding very high and low noise levels due to numerical instabilities. Like DreamFusion~\cite{poole_2023_dreamfusion}, we set $\omega = 100$ for classifier-free guidance and higher guidance weights give improved sample quality.
Intuitively, this loss perturbs $\tilde{\mathcal{S}}_a$ with a random amount of noise corresponding to the timestep $t$, and estimates an update direction that follows the score function of the diffusion model to move to a higher density region.
The ASDS loss function encourages the synthesized sketch to match the given text prompt, while also preserving the style and structure of the original sketch.
At each iteration, we backpropagate the loss through the differentiable rasterizer and update the control points and opacity of strokes directly until convergence of the loss function.

\noindent\textbf{Loss objectives of our final model.} To further enhance the quality of the synthesized sketches, we incorporate the JVSP and ASDS losses. Specifically, we first obtain the initial results with the JVSP loss, then we fine-tune the differentiable rasterizer together with the ASDS loss. We found such design can achieve the best performance. The ASDS loss predicts the gradient update direction directly, thus its loss value does not require weight balancing. In the JVSP loss, we set the weight of the LPIPS item to 0.2, and the weight of the CLIP visual item to 1. To compute the L2 distance between intermediate level activations of CLIP, we follow the CLIPasso~\cite{clipascene_vinker_2022} method and use layers 3 and 4 of the ResNet101 CLIP model.
As shown in Figure~\ref{fig:ablation}, we compare the performance of the vanilla version with that of the version using only the ASDS loss, and the proposed final version. Our experiments show that the final version improves both the generation quality and efficiency.

\subsection{Joint Attention-based Stroke Initialization}
\label{sec:stroke_init}
\begin{figure*}[t]
\vspace{-1em}
\centering
\includegraphics[width=1.0\textwidth]{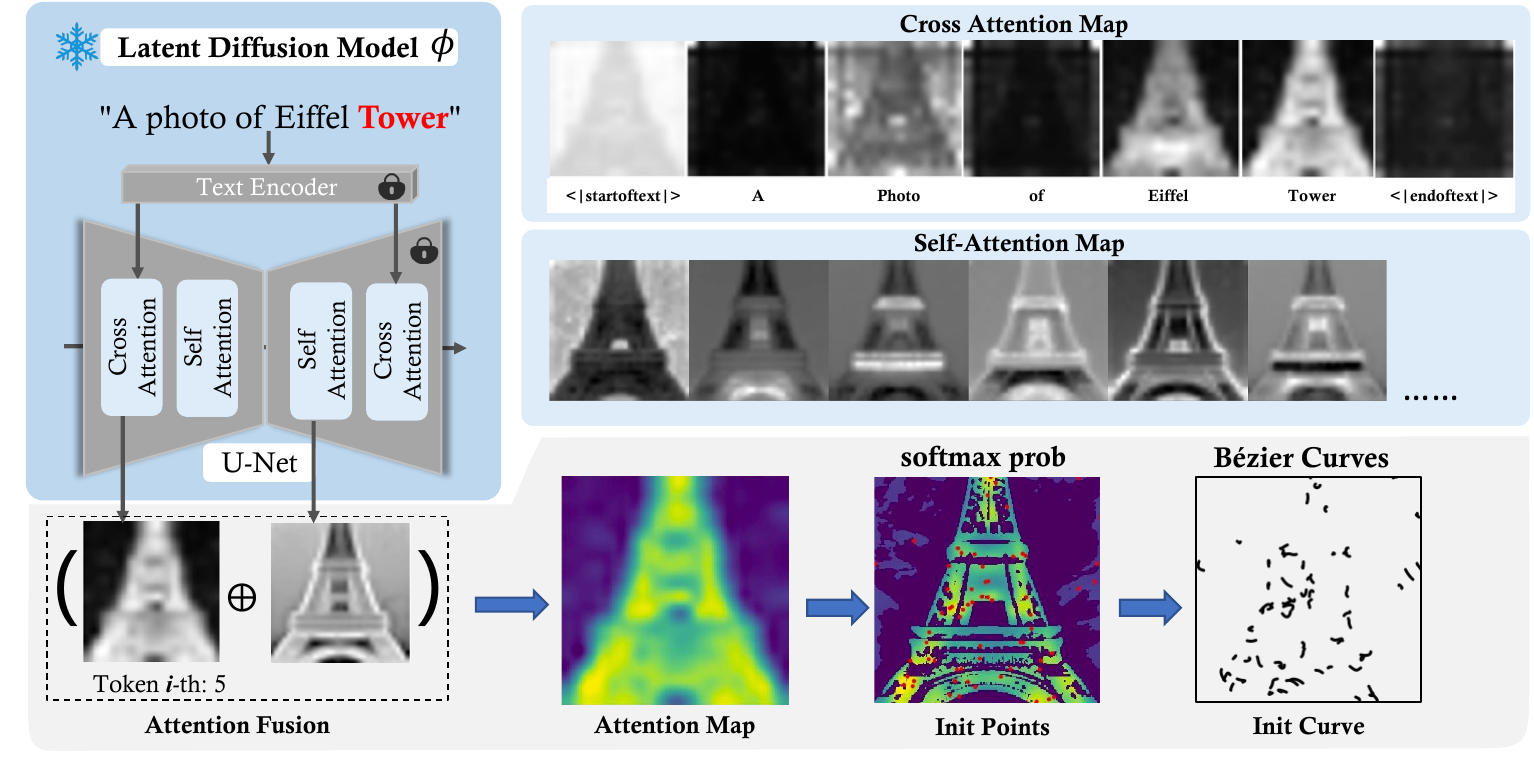}
\vspace{-2em}
\caption{
\textbf{Stroke Initialization. }
We utilize the U-Net of the frozen Latent Diffusion Model (LDM) to extract both self-attention and cross-attention maps. As shown in the dashed box, Attention Fusion is performed by taking the mean of the self-attention maps and combining it with the cross-attention map corresponding to a specific semantic token (\textit{e.g.}, the 5th token, "Tower"). The fused attention map is then processed into a probability distribution (softmax prob) to sample the initial points for Bézier curves. Note that token indexing starts at 1, as the 0-th index is reserved for the CLIP \texttt{<|startoftext|>} token.
}
\label{fig:attn_init}
\vspace{-1.5em}
\end{figure*}
The highly non-convex nature of the ASDS loss function makes the optimization process highly sensitive to initialization, especially in multi-instance scenarios where strokes must be carefully placed to capture the overall semantics of a free-hand sketch.
To improve convergence towards semantic depictions, we initialize the curve control points guided by the attention maps of a text-conditioned Latent Diffusion Model (LDM). 

As illustrated in Figure~\ref{fig:attn_init}, the U-Net within the LDM utilizes two types of attention mechanisms: self-attention and cross-attention. The structure and appearance of the LDM-generated image~\cite{ldm_2022_Rombach} rely on the interaction between spatial features and text embeddings during the diffusion process~\cite{prompt2prompt_hertz_2023}, which is primarily governed by the cross-attention layers~\cite{attention_vaswani_2017}. 
Visualizations indicate that cross-attention maps the spatial layout to specific words in the prompt, while self-attention dictates the overall geometric structure. 

Leveraging this observation, we linearly combine the probability distributions of these two attention maps. Specifically, we select the cross-attention map corresponding to a specific semantic token and fuse it with the averaged self-attention map. This fusion process is formalized as:
\begin{equation}
    \mathcal{A}_{\text{final}} = \lambda \mathcal{A}_{\text{cross}}^i + (1-\lambda) \bar{\mathcal{A}}_{\text{self}}
\end{equation}
where $\lambda$ is a weighting coefficient, $i$ denotes the index of the target token in the text prompt, and $\bar{\mathcal{A}}_{\text{self}}$ represents the mean of the self-attention maps.

Finally, we normalize this fused attention map using a softmax function to create a spatial probability distribution. We sample $n$ coordinates from this distribution to serve as the starting control point, $p_m^1$, for each of the $n$ Bézier curves. The remaining three control points ($p_m^2$, $p_m^3$, $p_m^4$) are initialized within a small local radius ($0.05$ of the image size) around $p_m^1$, defining the initial set of Bézier curves $\{ \{p_m^j\}_{j=1}^4 \}_{m=1}^n$. Empirical results demonstrate that our attention-fusion initialization significantly accelerates rendering and improves the final sketch quality compared to random initialization.

\section{Results}
\subsection{Qualitative Evaluation}

\begin{figure}[h]
\centering 
\includegraphics[width=\linewidth]{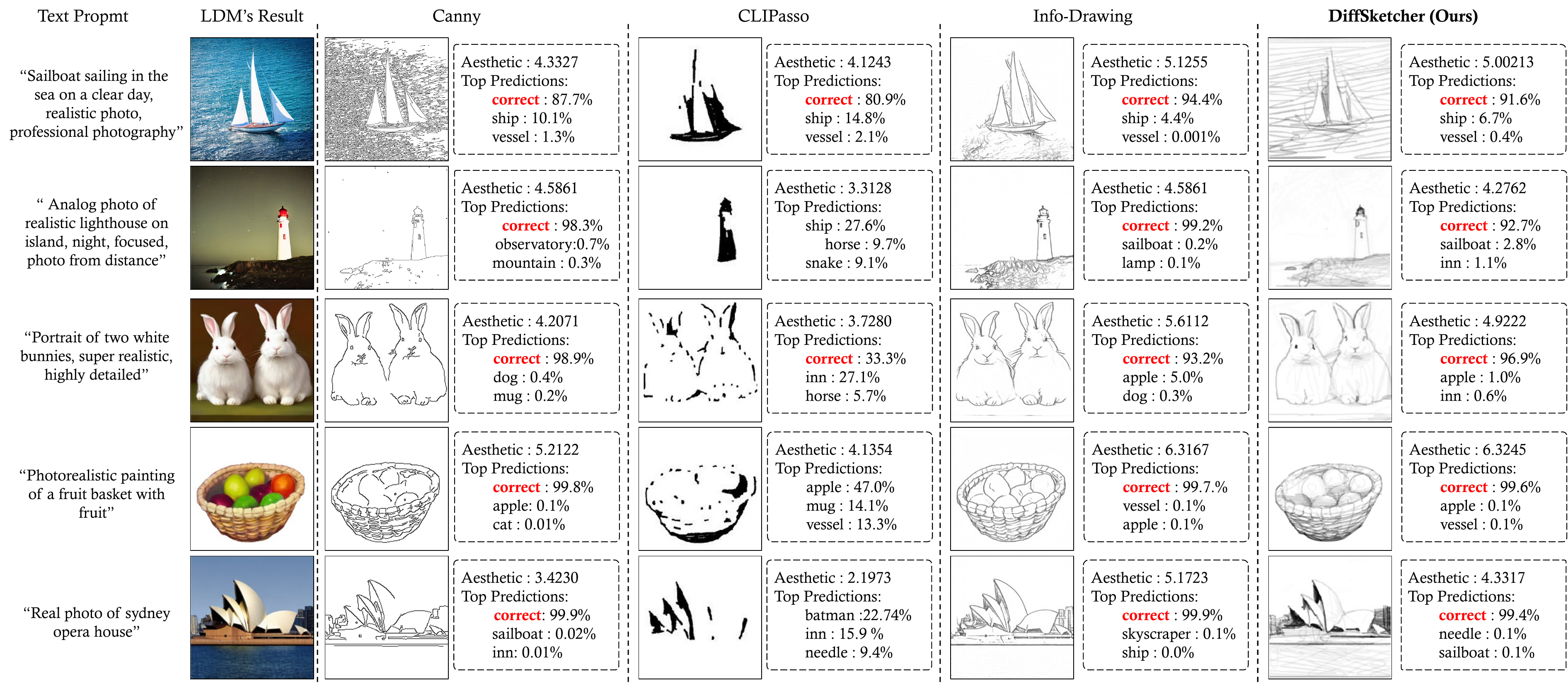}
\vspace{-1.5em}
\caption{\textbf{Comparison with existing sketch generation methods.} Our method produces sketches that are more semantically aligned with the input prompts and exhibit higher visual quality.}
\vspace{-0.5em}
\label{fig:compare}
\end{figure}

\label{sec:qualitative}
As shown in Figure~\ref{fig:more_results}, we demonstrate that our approach offers the ability to produce object-level and scene-level sketches based on a textual prompt, with the flexibility to manipulate the level of abstraction through the number of strokes. It is effective in generating accurate sketches regardless of prompt complexity, including simple or non-existent objects, iconic constructions, and detailed scenes. The utilization of the robust prior of stable diffusion allows for a favorable initialization, promoting the production of high-quality sketches with significantly fewer iterations.
Figure~\ref{fig:compare} qualitatively compared our DiffSketcher with those of  CLIP-based~\cite{vinker_2022_clipasso} and edge extraction-based methods~\cite{canny_1986}.
The Canny edge extraction algorithm extracts an excessive number of edges and produces untidy sketches, as observed in the first example in Fig.~\ref{fig:compare}. CLIPasso uses visual distance metrics to guide gradient-based optimization. On the task of drawing scene-level sketches, CLIPasso can only draw part of the foreground, and the background part is missing. 

\begin{figure}[h]
\centering 
\includegraphics[width=1.0\linewidth]{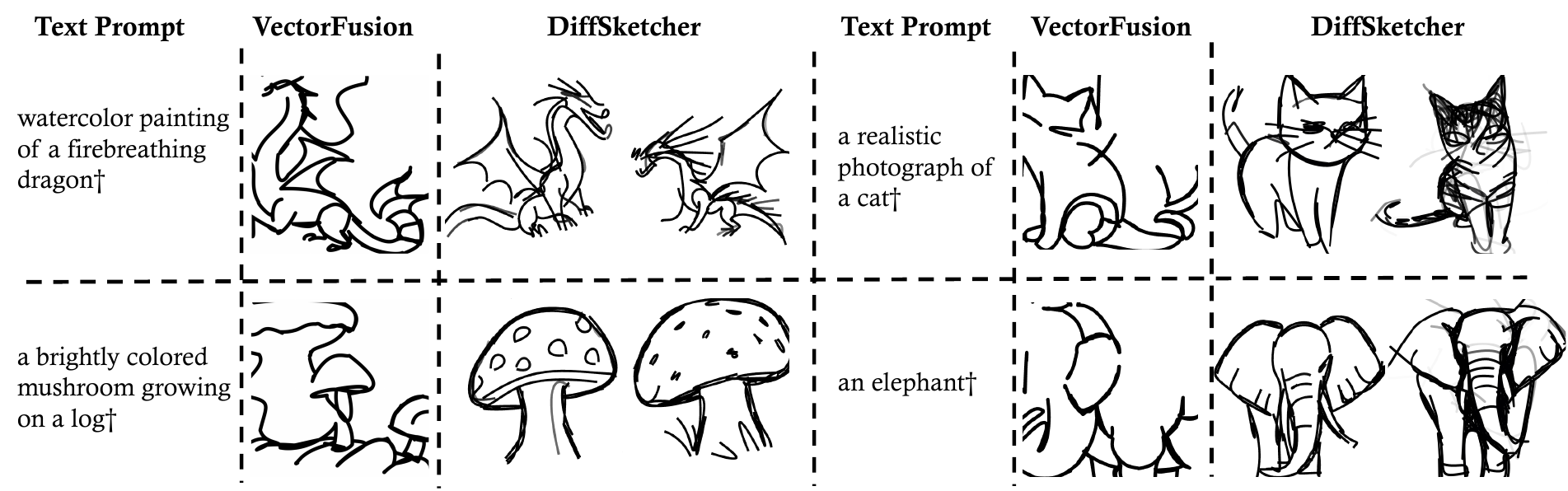}
\vspace{-1.5em}
\caption{
\textbf{Qualitative comparison with VectorFusion(VF)~\cite{jain2022vectorfusion}}.
VF's results are based on its original paper and project website, with a text prompt suffix of ``minimal 2D line drawing trending on ArtStation''. 
Note that our results were optimized from scratch using ASDS \textit{without} specifically designed text prompt suffix.
}
\vspace{-1em}
\label{fig:VF_sketch}
\end{figure}

We also compare our work with the recent work VectorFusion~\cite{jain2022vectorfusion}. The results are shown in Fig.~\ref{fig:VF_sketch}. This work is highly relevant to ours as it also explores the potential of diffusion model in generating vector graphics. However, there are notable differences between our appraoch and VectorFusion in terms of task setting, model design, and performance. Firstly, our method, DiffSketcher, primarily focuses on generating vector sketches based on text input. In contrast, VectorFusion aims to generate a broader range of vector graphics, including iconography and pixel art. It is important to highlight that our method has the capability to easily extend its functionality to generate other types of vector graphics \textit{without} changing primitives (as shown in Fig.~10 of Supp.~\ref{supp:compare2T2V}).
Secondly, DiffSketcher follows a distinct pipeline compared to VectorFusion. VectorFusion employs different model variants for different types of vector graphics. For generating vector sketches, our approach differs from VectorFusion mainly in two aspects: (1) \textbf{Initialization}: VectorFusion randomly initializes the rasterizer (i.e., the location of the control points), while our approach utilizes the attention layer of LDM, guided by text prompts, for initialization. This significantly improve both efficiency and synthesis quality. (2) \textbf{Optimization}: While VectorFusion solely optimizes the position of control points, our method also optimize the opacity of stroke. This contributes to enhance the visual quality of the synthesized sketches. As shown in Fig.~\ref{fig:VF_sketch}, our approach significantly outperforms VectorFusion in generating vector sketches, and achieves comparable performance in generating other types of vector graphics (see Fig.~10 of Supp.~\ref{supp:compare2T2V}). It is worth noting that our model variant shown in Fig.~\ref{fig:VF_sketch} utilizes only ASDS loss for optimization and initializes the strokes randomly. This was done to ensure a fair comparison with VectorFusion. Our full model is expected to deliver better performance.

\subsection{Quantitative Evaluation}
\label{sec:quantitative}
Evaluating text-to-sketch synthesis is challenging due to the absence of ground truth sketches. Therefore, we focus on three indicators: consistency between the generated sketch and text prompt, the aesthetic quality of the sketch, and the recognition accuracy of the sketch.
To measure the consistency between the generated sketch and input text, we calculate the mean of cosine similarity of CLIP embeddings for the generated sketches and the text captions used to generate them. Our method achieves a cosine similarity of $0.3494$, which is higher than the $0.328$ achieved by  Canny algorithm and the $0.3075$ achieved by CLIPasso.
Aesthetic appeal is subjective, and being visually appealing is a personal experience and preference. However, when describing the visual appeal of a hand-drawn sketch, various factors can be considered, such as line quality, texture, and style. As an evaluation of our proposed method, we use CLIP-based aesthetic indicators~\cite{aesthetic_predictor} to calculate the aesthetic value for samples of multiple categories. Figure~\ref{fig:compare} compares aesthetic values achieved by various methods. Our method achieves a mean value of $4.8206$ for the aesthetic value on a large number of examples, which is higher than the $4.3682$ achieved by  Canny and the $4.0821$ achieved by CLIPasso~\cite{vinker_2022_clipasso}.
\subsection{Ablation Study}
\label{sec:ablation}
\begin{figure}[h]
\centering 
\vspace{-0.5em}
\includegraphics[width=1\linewidth]{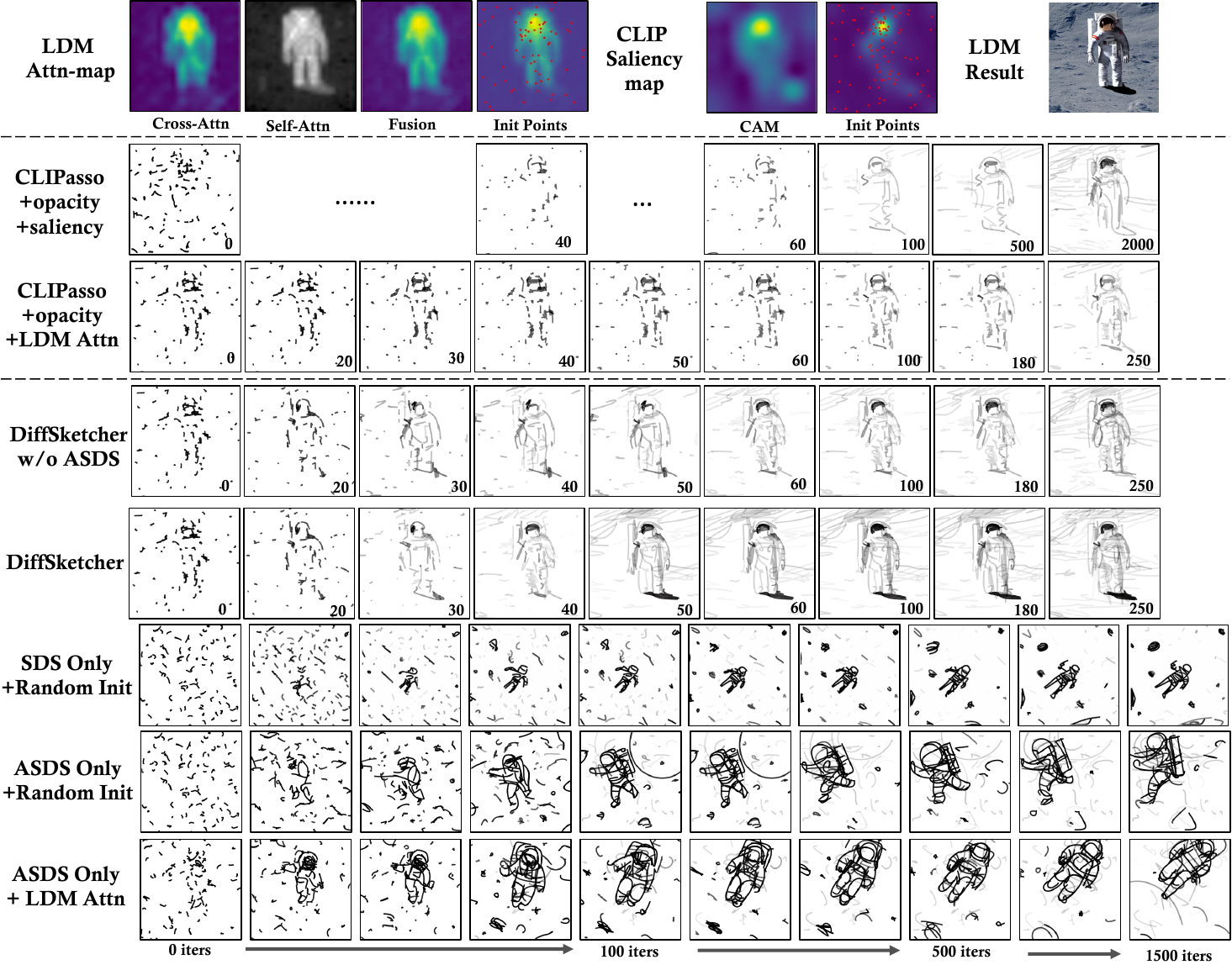}
\vspace{-1.5em}
\caption{
\textbf{Qualitative results of ablation study.}
\textbf{Top}: two initialization strategies used by DiffSketcher (\textit{i.e.}, LDM Attention) and CLIPasso~\cite{vinker_2022_clipasso} (\textit{i.e.}, CAM~\cite{gradcam_selvaraju_2017} Saliency map). A result of sampling from LDM . The \textbf{2nd} and \textbf{3rd} rows: Comparison of the two initialization strategies on the convergence speed. The \textbf{4th} and \textbf{5th} rows: The effect of JVSP (Section~\ref{sec:vanilla_version}) and ASDS loss (Section~\ref{sec:ASDS_loss}). 
The \textbf{5th} row shows the loss of SDS without data augmentation.
The \textbf{6th} and \textbf{7th} rows: ASDS loss leads to more diverse results when strokes are randomly initialized. Text prompt used in this example: ``Astronaut on Asteroid, galaxy background''.
}
\label{fig:ablation} 
\end{figure}
We conducted a series of experiments to demonstrate the effectiveness of the proposed initialization strategy and the effects of ASDS loss and JVSP loss, respectively. The top row of Fig.~\ref{fig:ablation} compares two initialization strategies used by DiffSketcher and CLIPasso~\cite {vinker_2022_clipasso}, namely LDM Attention and CAM~\cite {gradcam_selvaraju_2017} Saliency map, respectively. The text-driven diffusion model produces more precise attention map than the saliency map obtained by the CLIP-based method due to LDM's superior generation ability. The cross-attention feature of LDM can efficiently activate relevant regions based on token areas, while the self-attention layer can effectively differentiate foreground from background down to the pixel level. Through combining both mechanisms, the initialization area becomes more precise. As a result, superior quality sketches are produced with a reduced number of optimization steps, as illustrated in the second and third rows of Fig.~\ref{fig:ablation}. The proposed initialization strategy improves sampling quality and efficiency, which is critical for non-convex objective function optimization. 

In DiffSketcher, our JVSP loss consists of a CLIP loss and a LPIPS loss, and it is important to note that ASDS and JVSP do not conflict with each other. As illustrated in Fig.~\ref{fig:ablation}, the 4th and 8th rows highlight the effects of JVSP loss and ASDS loss, respectively. When only JVSP loss is used (4th row), the generated sample closely approximates the result of LDM. On the other hand, when only ASDS loss is employed (8th row), the generated sample aligns with the text prompt semantically, but does not follow the attention map of the LDM. Additionally, using ASDS loss results in more diverse sampling outcomes during the synthesis process. For instance, the location of the astronaut's head may change throughout the drawing process when ASDS loss is used, while it remains in the same location when the JVSP loss is employed. We utilize both JVSP loss and ASDS loss for optimization. As shown in the 5th row (labeled as "DiffSketcher"), the combination of these two losses leads to synthesized sketches with more intricate details and a visually more realistic appearance.

\section{Conclusion}
In this work, we present \textit{DiffSketcher}, a novel and effective approach that bridges the gap between natural language and free-hand sketch synthesis. By leveraging pretrained text-to-image diffusion models, our method generates diverse, high-quality vector sketches from text prompts without requiring large-scale sketch datasets or text-sketch pairs. Furthermore, we provide a comprehensive analysis of key design choices---including stroke initialization strategies, loss functions, and differentiable rasterization properties---offering valuable insights for future research. Ultimately, \textit{DiffSketcher} demonstrates significant potential as a versatile tool for downstream applications in design and education.

\noindent\textbf{Limitations.} \quad
Our approach currently has two main limitations. First, there is no explicit correlation between the semantic complexity of the text prompt and the abstractness of the resulting sketch, which can occasionally lead to suboptimal representations. A promising direction to address this is establishing an adaptive mapping between prompt complexity and the allocated stroke count (see Supp.~\ref{supp:failure_case} for details). Second, the stylistic diversity of the generated sketches is somewhat constrained. Future work could alleviate this by integrating a style transfer module directly into the synthesis pipeline.

\section{Acknowledgement}
This work is supported by the National Natural Science Foundation of China (No. 62002012, No. 62006012, and No. 62132001) and CCF-Baidu Open Fund. It is also supported in part by The Hong Kong Jockey Club Charities Trust under Grant 2022-0174, in part by the Hong Kong Research Grant Council under General Research Fund (17203023), in part by the Startup Funding and the Seed Funding for Basic Research for New Staff from The University of Hong Kong, and in part by the funding from UBTECH Robotics. 

\medskip

\newpage
\clearpage

\bibliography{bibref}
\bibliographystyle{plain}


\newpage
\clearpage

\appendix

\section*{Supplementary}
\section*{Overview}
This supplementary material is organized into several sections that provide additional details and analysis related to our work on DiffSketcher. Specifically, it will cover the following topics:
\begin{itemize}
\item In section~\ref{supp:implement}, we provide the implementation details of DiffSketcher.
\item In section~\ref{supp:compare2T2V}, we present a qualitative comparison of our DiffSketcher with another two text-to-SVG methods, CLIPDraw~\cite{clipdraw_frans_2022} and VectorFusion~\cite{jain2022vectorfusion}. We compare results generated by these methods and analyze the differences in terms of visual quality and semantic consistency.
\item In section~\ref{supp:compare2LDM}, we compare sketches generated by our DiffSketcher with those directly sampled from the LDM (\textit{i.e.}, Stable Diffusion~\cite{ldm_2022_Rombach}) and analyze the differences in their style.
\item In section~\ref{supp:user_study}, we conducted a perceptual study to assess the authenticity of the synthesized sketches.
\item In section~\ref{supp:init}, we compare the results of three different strategies for stroke initialization.
\item In section~\ref{supp:track}, we visualize how DiffSketcher gradually sketches an object or a scene. 
\item In section~\ref{supp:stroke_width}, we provide example sketches with different stroke widths.
\item In section~\ref{supp:metric}, we introduce the details of the evaluation metrics used in our experiments.
\item In section~\ref{supp:failure_case}, we show several examples of failure cases.
\end{itemize}

\section{Implementation Details of DiffSketcher.}
\label{supp:implement}
We begin by describing the vanilla version of our approach (Section~\ref{sec:vanilla_version}), which involves sampling an image from the latent diffusion model \cite{ldm_2022_Rombach} and then automatically sketching it using DiffSketcher. Specifically, given a text prompt, we use a DDIM solver \cite{ddim_song_2021} to sample a raster image from the latent diffusion model in 100 steps with classifier-free guidance \cite{classifierfree_2022_ho}, using a scale of $\omega = 7.5$. To apply the Augmentation SDS loss (Section~\ref{sec:ASDS_loss}), we sample a noise level $t$ from the uniform distribution $\mathcal{U}(0.05, 0.95)$, avoiding very high and low values that can cause numerical instabilities. For classifier-free guidance, we set $\omega = 100$, as we found that a higher guidance weight leads to better sample quality. This is larger than the scale used in image sampling methods, which is likely due to the mode-seeking nature of our objective, leading to over-smoothing at small guidance weights.

In DiffSketcher, the number of strokes $n$ is defined by the user, and we use $4$ duplicates for image augmentation to maintain recognizability under various distortions. For this purpose, we apply the \textit{torch.transforms.RandomPerspective}, \textit{torch.transforms.RandomResizedCrop}, and \textit{torch.transforms.RandomAdjustSharpness} functions in sequence. It is worth noting that data augmentation is not the focus of our work, and experimental results show that the choice of augmentation strategies does not affect results significantly.

In Section~\ref{sec:method}, we define a sketch as a set of $n$ strokes $\{s_1, \dots, s_n\}$ with the opacity attribute placed on a white background. To optimize the control points and opacity, we use two Adam optimizers. Specifically, we set the learning rate of the control point optimizer to 1.0 and the color optimizer to 0.1.

\section{Comparison to Existing Text-to-SVG Work.}
\label{supp:compare2T2V}
\begin{figure}[th]
\centering 
\includegraphics[width=1.0\linewidth]{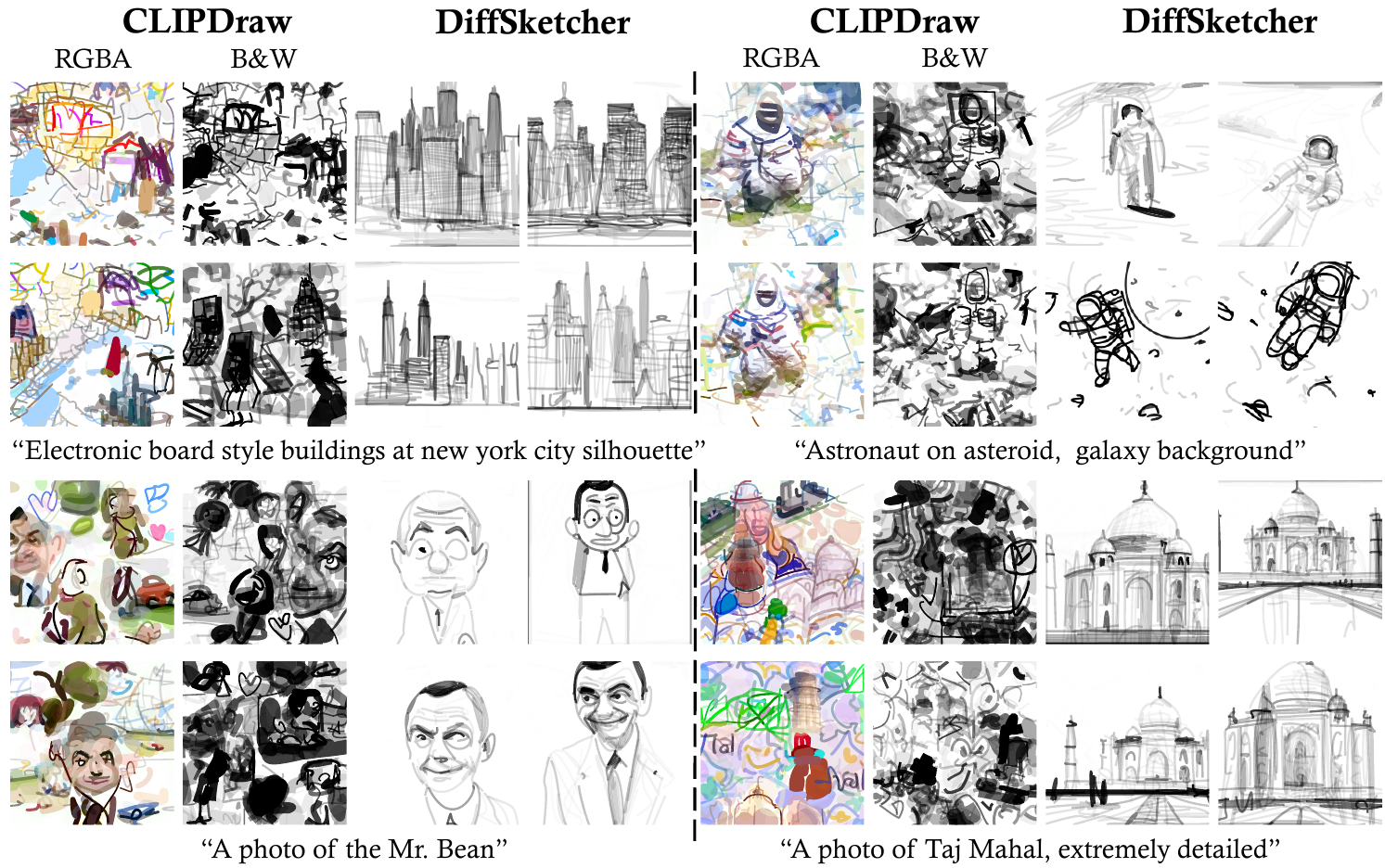}
\vspace{-1em}
\caption{
Comparison of the results synthesized by CLIPDraw and DiffSketcher. Specifically, for each example, we compare the results generated by CLIPDraw (Left) and our DiffSketcher (Right) given the same text prompt. We implemented two versions of CLIPDraw: the RGBA version (original version) and the B\&W version. The B\&W version forces the stroke color to be black to mimic the sketch style. Our results are visually more pleasing and meaningful.
}
\label{fig:compare_text2svg}
\end{figure}
This section presents a qualitative comparison between DiffSketcher and CLIPDraw~\cite{clipdraw_frans_2022}. CLIPDraw is a CLIP-based method introduced for text-to-SVG generation. It gradually optimizes the position and colors of the curves by the gradient descents computed by comparing the cosine similarity of the text prompt and the generated drawings. Our method differs from CLIPDraw in two ways, one is the stroke initialization, and more importantly, ours is equipped with the Augmentation SDS (ASDS) loss. As illustrated in Fig.~\ref{fig:compare_text2svg}, CLIPDraw struggles to synthesize a meaningful and visually pleasing drawing, no matter whether with colors or not. It can be explained by the fact that the CLIP model is not a generative model, and it can only provide guidance from a highly-semantic perspective. In contrast, our DiffSketcher can generate sketches that are semantically consistent with the input text, and exhibit high aesthetic quality. This is because the proposed ASDS loss can distill the drawing capability of the latent diffusion model (LDM) into the differentiable rasterizer. These results also suggest the effectiveness of the ASDS loss and the benefits of leveraging the power of the LDM. VectorFusion~\cite{jain2022vectorfusion} is highly relevant to our work but it aims to produce general vector graphics, such as iconography. Although our proposed method is designed to generate vector sketches, it can easily extend to generate other types of vector graphics. In Fig.~10, we compare results obtained by VectorFusion and our DiffSketcher. 


\begin{figure}[t]
\centering 
\includegraphics[width=0.98\linewidth]{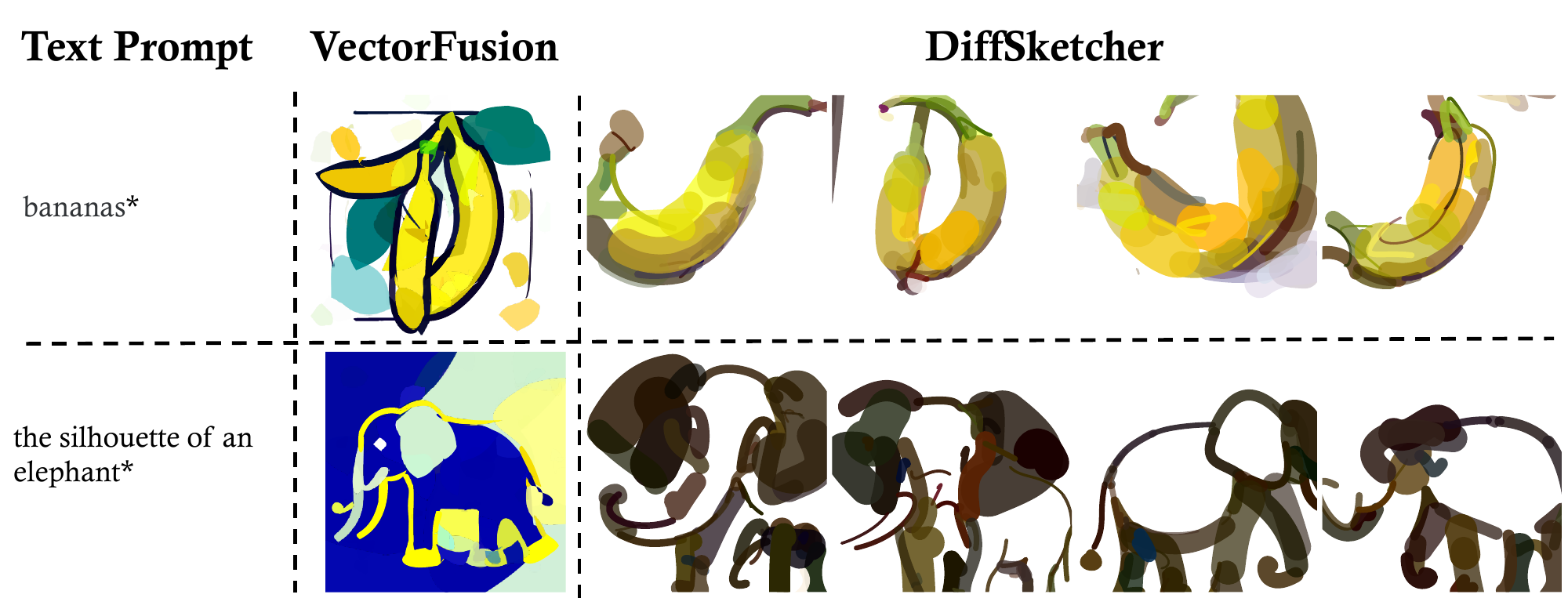}
\vspace{-1em}
\caption{
Qualitative comparison with VectorFusion(VF) ~\cite{jain2022vectorfusion}. VF's results were copied from \textit{Figure~2} of its original paper, with a text prompt suffix of ``minimal 2D line drawing trending on ArtStation''. In contrast, our results were optimized from scratch using ASDS without specifically designed text prompt suffix.
}
\vspace{-1em}
\label{fig:VF_vs_ours}
\end{figure}

\section{Comparison to Existing Text-to-Image Work.}
\label{supp:compare2LDM}
\begin{figure}[t!]
\centering 
\includegraphics[width=1\linewidth]{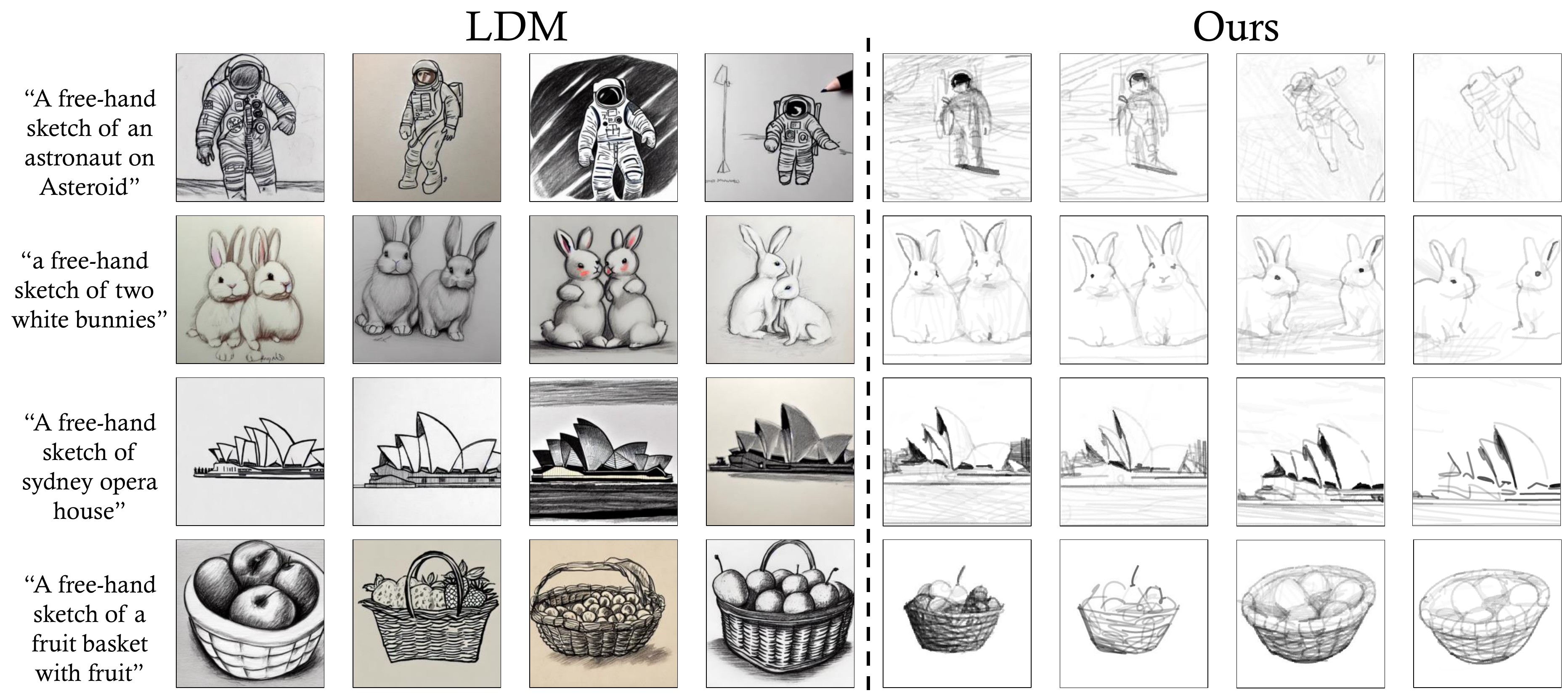}
\vspace{-1em}
\caption{
Comparison of sketches generated by sampling from the LDM using the specified text prompt (Left) and ours (Right). 
}
\label{fig:ldm_result}
\end{figure}

\begin{figure}[t!]
\centering 
\includegraphics[width=0.75\linewidth]{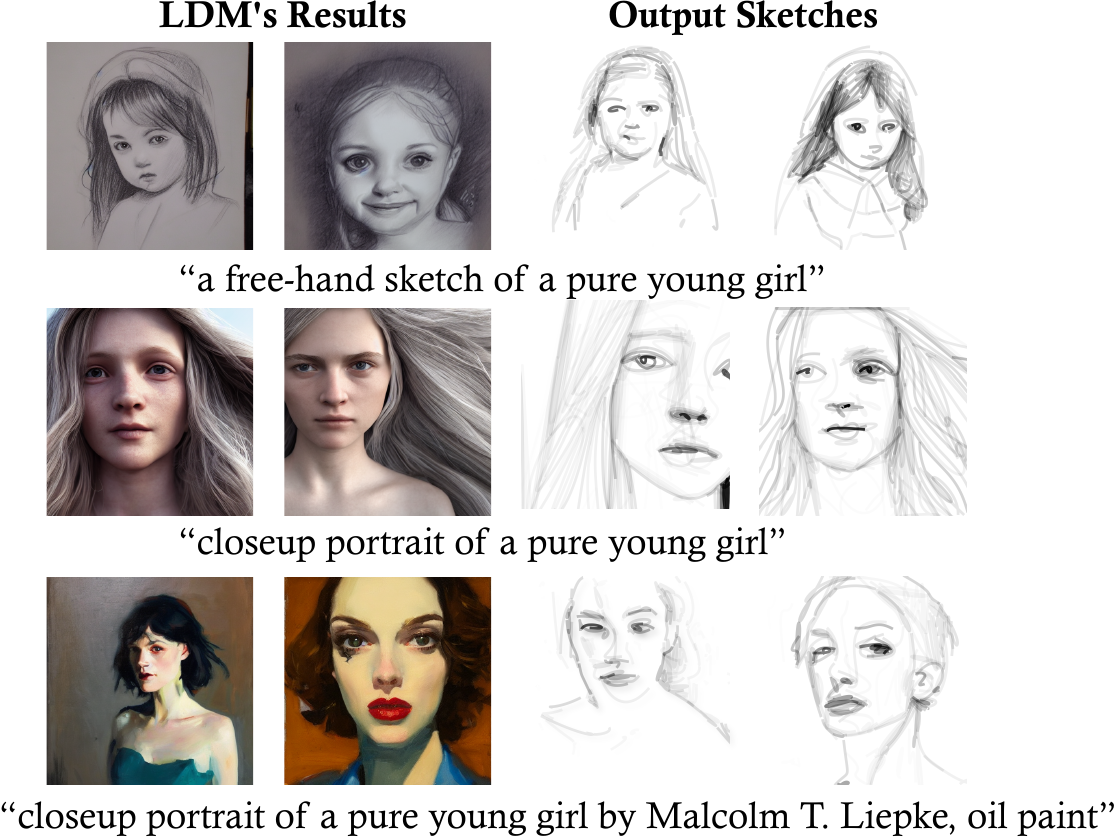}
\vspace{-0.5em}
\caption{
The style of the generated sketches is not significantly affected by the keywords used in the text prompt.
}
\label{fig:prompt_compare}
\end{figure}
In this section, we compare the sketches directly generated by LDM using specific text prompts. To encourage the results to be abstract and follow the free-hand sketch style, we append a suffix to the text prompt: "A free-hand sketch of xxx on a white background, trending on ArtStation. Keep abstract." This prompt was tuned qualitatively to capture the desired style and artistic expression.

The results are shown in Fig.~\ref{fig:ldm_result}. It is clear that the LDM is capable of generating high-quality \textit{raster} sketches in a free-hand sketch style. However, different from the \textit{vector} sketches generated by our DiffSketcher, LDM's sketches have two distinguished characteristics: 1. they are more delicate and like professional sketches. 2. their background exhibits the paper texture. We suppose this is because most sketches used for training LDM are photographs of professional sketches drawn on paper. By contrast, the sketches synthesized by our DiffSketcher are more like the style of free-hand sketch and exhibit different levels of abstractness.

It is worth noting that our method does not require indicating "free-hand sketch" in the text prompt. The drawing engine of our model is the rasterizer and it can naturally capture the sketch style. We also conduct experiments to show the results of using different keywords in text prompts, such as "sketch", "drawing", and "photo". The results are shown in Fig.~\ref{fig:prompt_compare}. We can see that although the style of the output images is different, the style of the generated sketches is not significantly affected by the keywords.
\section{User Study.}
\label{supp:user_study}

\begin{figure}[t!]
\centering 
\includegraphics[width=0.7\linewidth]{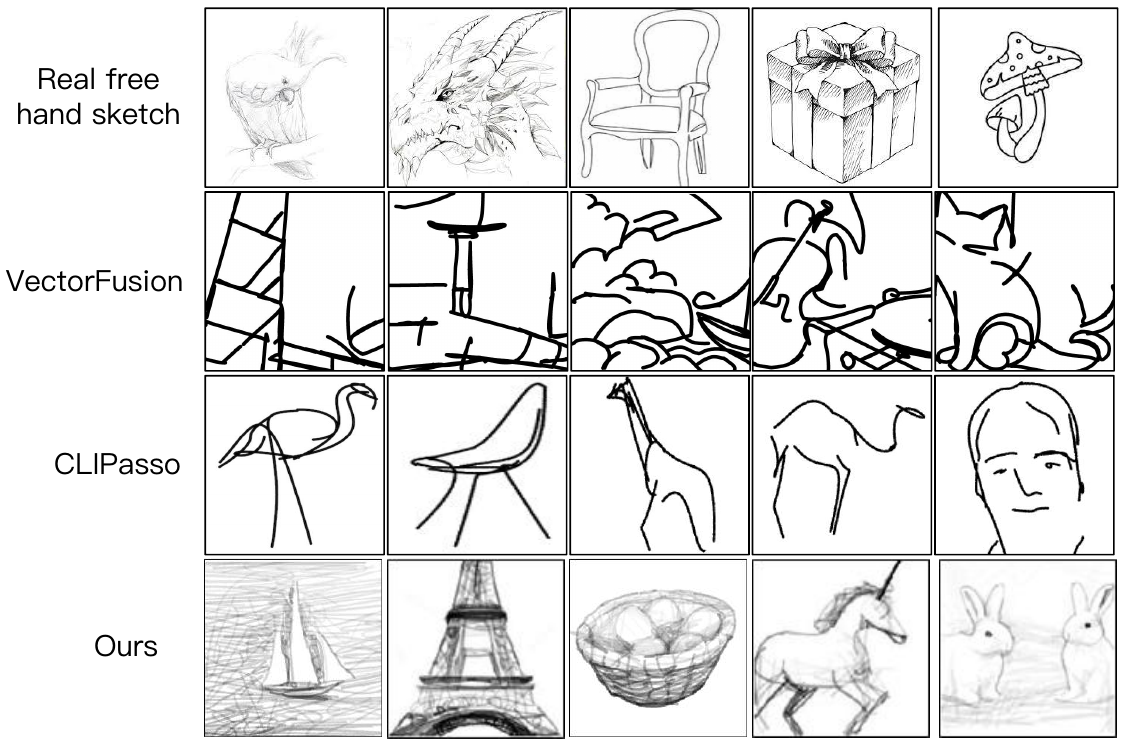}
\vspace{-1em}
\caption{
Partial sample visualization for conducting user research. The hand-drawn sketches were sourced from the Google. CLIPasso's and our results were sourced from respective paper, VectorFusion's results were sourced from their project homepage.
}
\label{fig:user_study}
\end{figure}

\begin{table}[h]
\centering
\caption{Results of the User Study. The Confusion score of real sketch means only 67\% real sketches are recognized as real.}
\vspace{-0.5em}
\label{tab:user_study}
\resizebox{34em}{!}{
\begin{tabular}{l|cccc}
\toprule
Metric / Method & CLIPasso~\cite{vinker_2022_clipasso} & VectorFusion~\cite{jain2022vectorfusion} & DiffSketcher (Ours) & Human Sketch \\
\midrule
Confusion Score & 0.39 & 0.33 & 0.65 & 0.67 \\
\bottomrule
\end{tabular}
}
\end{table}

To assess the authenticity of the synthesized sketches, we conducted a perceptual study. Specifically, We gathered a total of 90 synthesized sketches using three different methods (30 samples per method) and obtained 30 real sketches from Google Image by searching for "free-hand sketch". Figure~\ref{fig:user_study} shows a partial sample. We then mixed the real and fake sketches and distributed questionaires to 41 participants. The participants were asked to determine whether each sketch was drawn by a human or not, without any knowledge of its source. We utilized the confusion score as the evaluation metric, where a higher score indicates a greater likelihood of the generated sketches being recognized as real. The results are presented in Table~\ref{tab:user_study}. It is clear to see that our method produced sketches that were more frequently identified as real, highlighting the superior quality of our synthesized sketches.

\section{Stroke Initialization.}
\label{supp:init}
The highly non-convex nature of the ASDS loss and JVSP loss function makes the optimization process susceptible to stroke initialization, especially for generating scene-level sketches with multi-instances. To address this issue, we explore different stroke initialization strategies (as mentioned in Section~\ref{sec:ablation} and Fig.~\ref{fig:ablation}) and evaluate their impact on the performance of our model.
\begin{figure}[h]
\centering 
\includegraphics[width=0.95\linewidth]{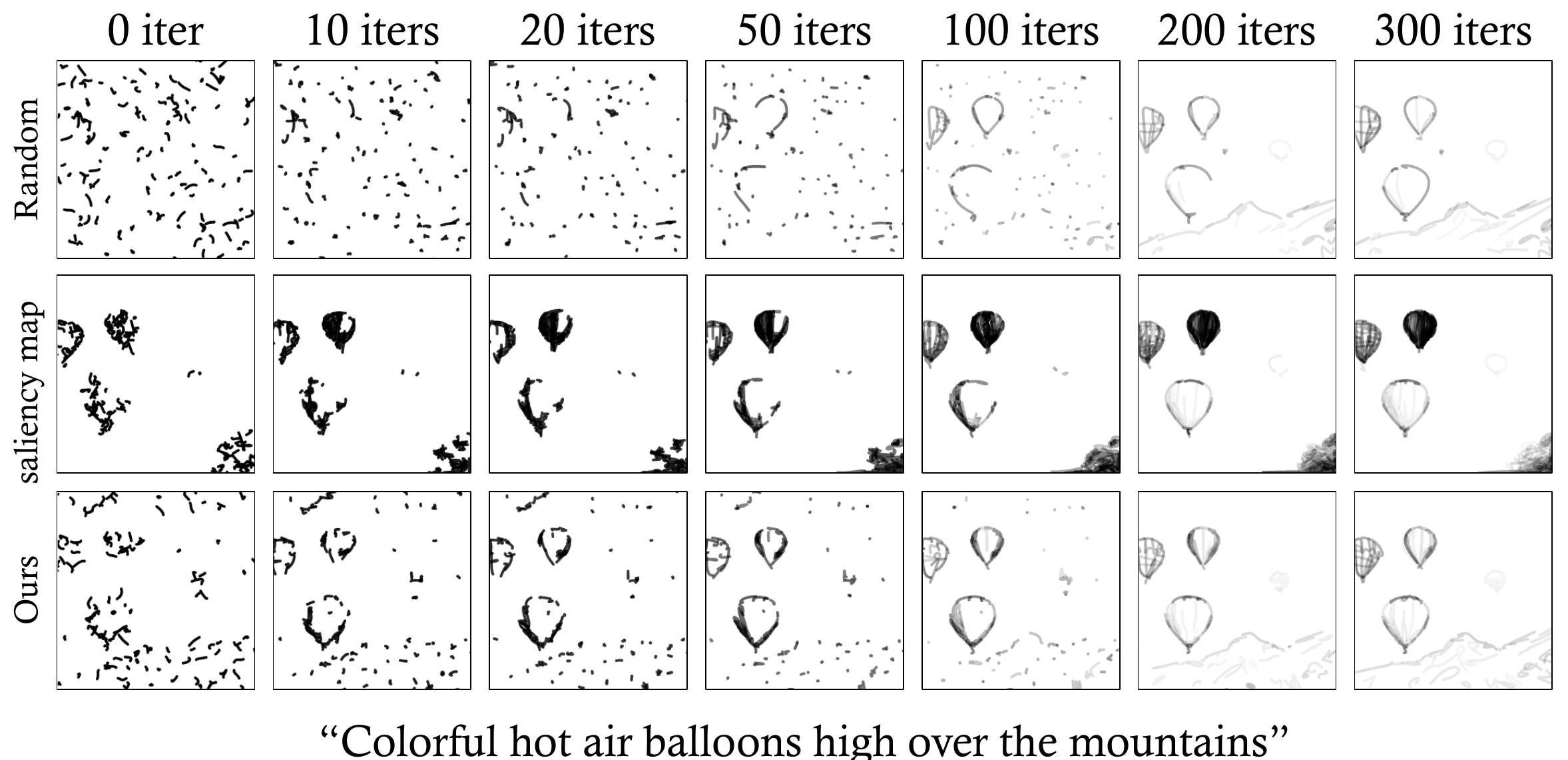}
\caption{
Comparison of the (intermediate) results when using different stroke initialization strategies. From top to bottom: (a) random initialization, (b) initialization based on the CLIP saliency map, (c) our proposed fusion attention.
}
\label{fig:stroke_init}
\end{figure}

As shown in Figure.~\ref{fig:stroke_init}, we compare three different stroke initialization methods: random initialization, initialization based on the CLIP saliency map~\cite{vinker_2022_clipasso}, and initialization based on our proposed fusion attention (ours). The experimental results demonstrate that our proposed strategy, initialization based on fusion attention, outperforms the other methods in terms of visual quality and synthesis efficiency. This initialization method can utilize the joint semantic and structural information from the input text and image (i.e., LDM results) to guide the stroke placement, resulting in more semantically meaningful and artistically expressive sketches. However, using a CLIP saliency map for initialization 
leads to a sketch with only salient objects while ignoring the background. Random initialization takes more iterations than ours to synthesize visually pleasing results.

\section{Visualization of Sketching Process.}
\label{supp:track}
\begin{figure}[h]
\centering 
\includegraphics[width=1.0\linewidth]{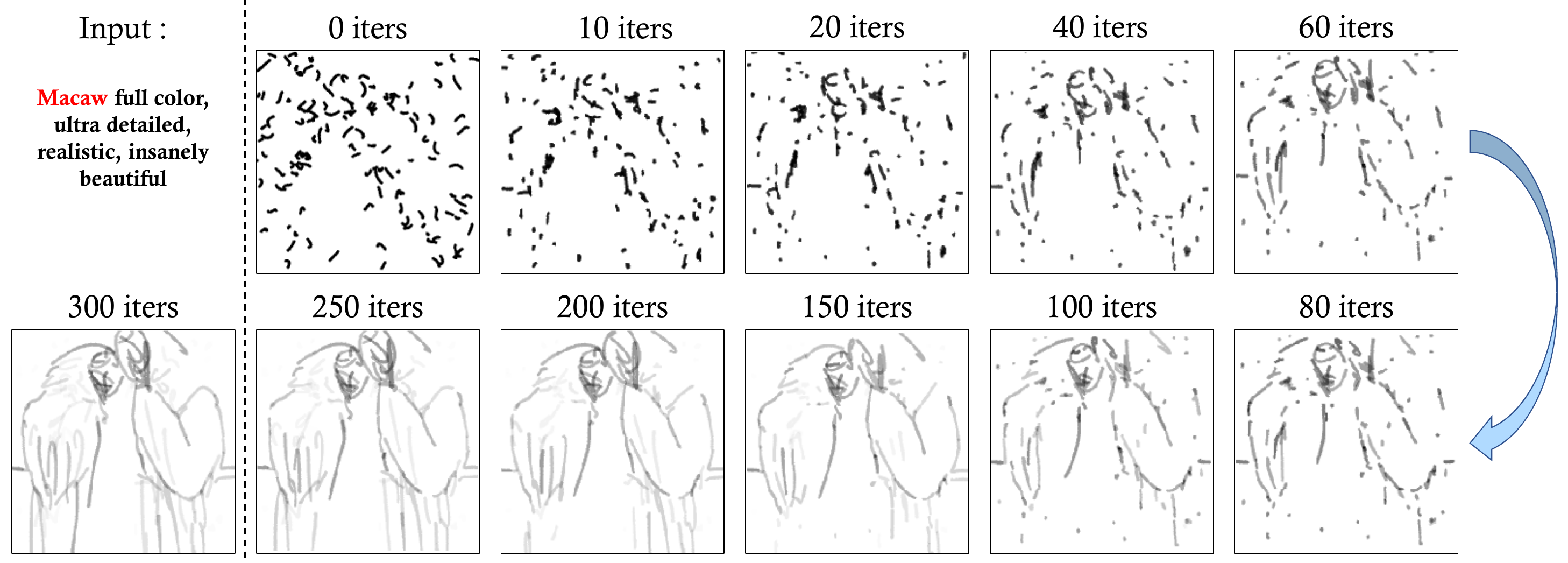}
\vspace{-1em}
\caption{
The intermediate results throughout the optimization process.
}
\label{fig:painting_process}
\end{figure}

In this section, we show the trace of 300 iters of sketching.
By visualizing the intermediate outputs during the generation process, we can gain insights into how our model sketches an object. Specifically, as shown in Fig.~\ref{fig:painting_process}, we can observe how the strokes are placed and refined over time to gradually form the desired sketch.

\begin{figure}[h]
\centering 
\includegraphics[width=0.85\linewidth]{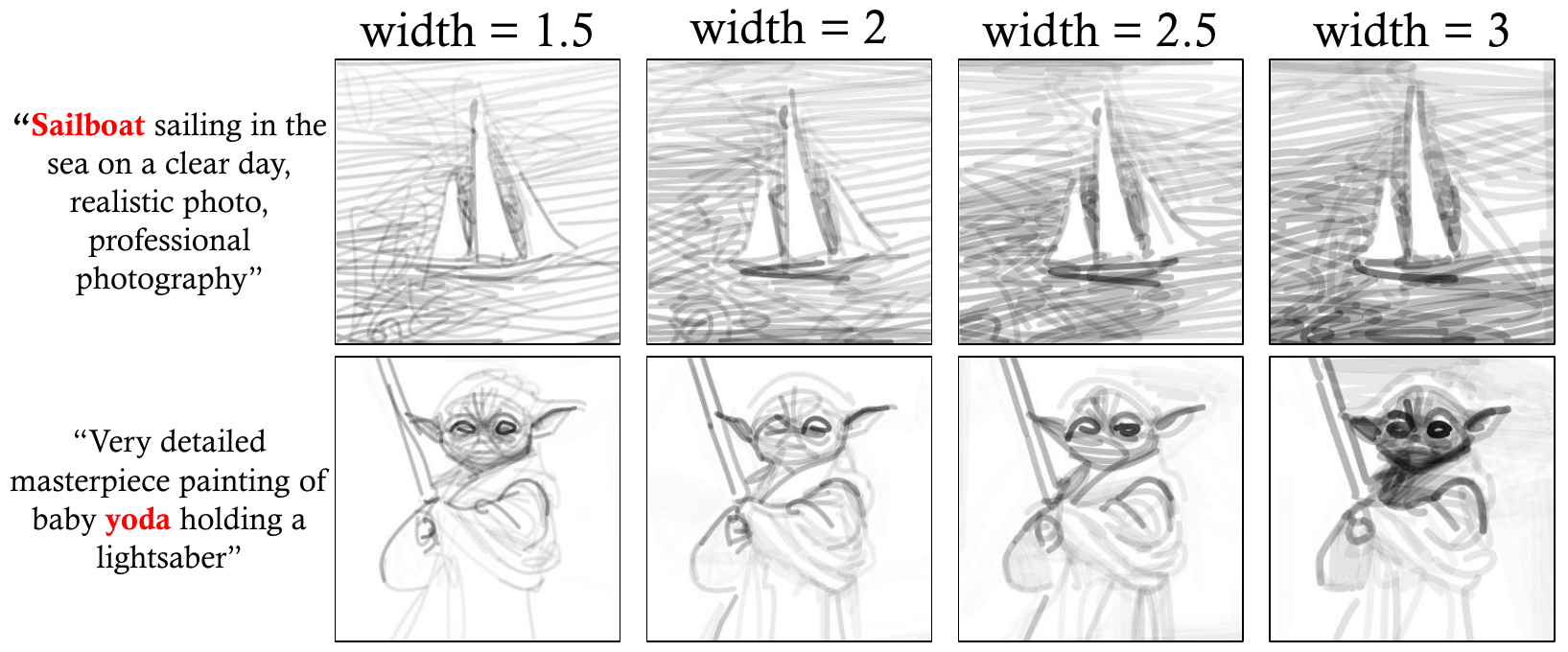}
\vspace{-0.5em}
\caption{
Different widths of the curves. The width increases from left to right.
}
\label{fig:stroke_widths}
\end{figure}

\begin{figure}[h]
\centering 
\includegraphics[width=0.65\linewidth]{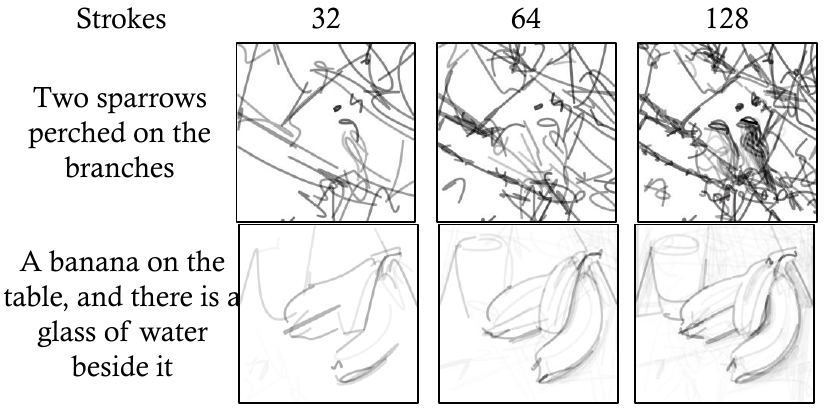}
\vspace{-0.5em}
\caption{
The failure cases.
}
\label{fig:failure_case}
\end{figure}

\section{Effect of the Stroke Width.}
\label{supp:stroke_width}
In Fig.~\ref{fig:stroke_widths}, we compare the results with different stroke widths given the same text prompt. In our implementation, we use a fixed stroke width for all strokes. Such a design is to simplify the optimization, making it computationally more efficient and less prone to overfitting. It can easily extend to include stroke width as a parameter for optimization, like ~\cite{clipdraw_frans_2022,Styleclipdraw_schaldenbr_2021}. 

\section{Evaluation Metrics.}
\label{supp:metric}
Evaluating text-to-sketch synthesis is challenging due to the absence of ground truth sketches. As we mentioned in Section~\ref{sec:quantitative} and Fig.~\ref{fig:compare}, we evaluate the models from three aspects: semantic consistency between the generated sketch and text prompt, the aesthetic quality of the sketch, and the recognizability of the sketch.

\textbf{Semantic Consistency Between the Generated Sketch and Text Prompt. } To measure the semantic consistency, namely the CLIP score~\cite{ldm_2022_Rombach,clipdraw_frans_2022}, we calculate the cosine similarity of CLIP ViT-L-14 embeddings of the generated sketches and corresponding input text prompts. 
Our method achieves a cosine similarity of $0.3494$, which is higher than  Canny~\cite{canny_1986} algorithm ($0.328$) and CLIPasso~\cite{vinker_2022_clipasso} ($0.3075$).

\textbf{The Aesthetic Quality of the Sketch. } To measure the aesthetic quality of generated sketches, we adopt the CLIP-based aesthetic indicator~\cite{aesthetic_predictor}. 
This indicator~\cite{aesthetic_predictor} consists of a CLIP ViT-L-14 backbone and a multi-layer perception (MLP), which is pre-trained on LAION~\cite{laion_schuhmann_2022} data.
Figure~5 of the main paper compares the aesthetic score of several examples generated by different methods. Sketches generated by our method obtain the highest scores.

\textbf{The Recognizability of the Sketch. } Finally, to measure the recognizability of the generated sketches, we use the CLIP ViT-L-14 model~\cite{clip_radford_2021} for zero-shot classification. Specifically, we first generate sketches for 34 categories\footnote{The 34 categories include "astronaut", "vessel", "observatory", "needle", "outer space", "earth", "iron man", "batman", "apple", "sailboat", "ship", "bunny", "castle", "cabin", "inn", "bike", "cat", "dog", "dragon", "snake", "horse", "fruit basket", "Sydney opera house", "lamp", "lighthouse", "mug", "desk", "macaw", "mountain", "river", "eiffel tower", "unicorn", "yoda", "skyscraper".}.
Then, we use the CLIP model to classify these sketches. In Fig.~\ref{fig:compare} of the main paper, we list the probabilities of the sketch being categorized into different classes. Since a Canny edge can preserve the object's contours and fine details, it achieves high recognizability. Compare our DiffSketcher with CLIPasso, our DiffSketcher can draw a more complete sketch. Therefore, our sketches are easier for CLIP to recognize.

\section{Failure Cases}
\label{supp:failure_case}
As shown in Fig.~\ref{fig:failure_case}, our approach mainly has two limitations. Specifically, one limitation is the lack of correlation between the text prompt and sketch abstractness. For instance, if the text prompt describes multiple objects but the number of strokes is set too small, the resulting sketches may be unsatisfactory. A possible solution is to estabilish a link between the complexity of the text prompt (such as the number of described objects) and the number of strokes to be used.

\end{document}